\documentclass{article}

\usepackage[table,x11names]{xcolor}  
\usepackage{colortbl}               
\usepackage{graphicx}             
\usepackage{booktabs}             
\usepackage{array}

\definecolor{firstplace}{RGB}{255, 255, 153} 
\definecolor{secondplace}{RGB}{200, 230, 240}

\usepackage{microtype}
\usepackage{makecell}
\usepackage{multirow}
\usepackage{subfigure} 
\usepackage{booktabs}
\usepackage{wrapfig}
\usepackage{extarrows}
\usepackage{colortbl}
\usepackage{tikz}
\usepackage{tablefootnote}
\usepackage{bbding}
\usepackage{pifont}
\usepackage{adjustbox}
\usepackage{bold-extra}
\usepackage{graphicx}
\usepackage{subfigure}
\usepackage{overpic}
\usepackage{float}
\usepackage{longtable}
\usepackage{ragged2e}
\usepackage{hyperref}

\usepackage{algorithm}
\usepackage{algorithmic}

\numberwithin{algorithm}{section}

\usepackage{amsmath}
\usepackage{amssymb}
\usepackage{mathtools}
\usepackage{amsthm}

\usepackage{graphicx}
\usepackage{booktabs}
\usepackage{caption}
\usepackage{tabularx}

\newtheorem{theorem}{Theorem}

\newtheorem{lemma}{Lemma}

\newtheorem{definition}{Definition}

\newtheorem{assumption}{Assumption}

\newtheorem{myrule}{Rule}

\usepackage[capitalize,noabbrev]{cleveref}

\Crefname{theorem}{Theorem}{Theorems}

\Crefname{proposition}{Proposition}{Propositions}

\Crefname{lemma}{Lemma}{Lemmas}

\Crefname{corollary}{Corollary}{Corollaries}

\Crefname{definition}{Definition}{Definitions}

\Crefname{myrule}{Rule}{Rule}

\Crefname{example}{Example}{Examples}

\Crefname{assumption}{Assumption}{Assumptions}

\Crefname{remark}{Remark}{Remarks}

\usepackage{titletoc}
\newcommand\DoToC{%
	\startcontents
	{\color{black} %
		\printcontents{}{1}{%
			\color{black}\textbf{Contents of Appendix}\vskip3pt\hrule\vskip5pt%
		}%
		\vskip3pt\hrule\vskip5pt}%
}

\usepackage[preprint]{2025}
\usepackage{floatflt}

\usepackage{comment}
\usepackage[utf8]{inputenc}
\usepackage[T1]{fontenc}
\usepackage{hyperref}
\usepackage{url}
\usepackage{booktabs}
\usepackage{amsfonts}
\usepackage{nicefrac}
\usepackage{microtype}

\usepackage[normalem]{ulem}  
\title{SamGoG: A Sampling-Based Graph-of-Graphs Framework for Imbalanced Graph Classification}

\hypersetup{
	colorlinks=true,
	linkcolor=black, 
	citecolor=green,
	filecolor=magenta,      
	urlcolor=magenta,
}

\author{
	\vspace{-25pt}\\
	\textbf{Shangyou Wang$^{1, 3, \dag}$,\quad Zezhong Ding$^{1, 3, \dag}$,\quad Xike Xie$^{2, 3}$\thanks{Corresponding Author \quad  $^\dag$Equal Contribution}}\vspace{3pt} \\
	{\small $^1$School of Artificial Intelligence and Data Science, University of Science and Technology of China (USTC})\\
	{\small $^2$School of Biomedical Engineering, USTC} \\
	{\small $^3$Data Darkness Lab, Suzhou Institute for Advanced Research, USTC}\\
	\texttt{\small \{wash\_you,zezhongding\}@mail.ustc.edu.cn, xkxie@ustc.edu.cn} 
	\vspace{-10pt}\\
}

\begin{document}

	\maketitle
	
	\vspace{-0.7em}
	\begin{abstract}
		\vspace{-0.7em}
		\emph{Graph Neural Networks} (GNNs) have shown remarkable success in graph classification tasks by capturing both structural and feature-based representations. However, real-world graphs often exhibit two critical forms of imbalance: class imbalance and graph size imbalance. These imbalances can bias the learning process and degrade model performance. Existing methods typically address only one type of imbalance or incur high computational costs. In this work, we propose SamGoG, a \underline{\bf sam}pling-based \emph{Graph-of-Graphs} (\underline{\bf GoG}) learning framework that effectively mitigates both class and graph size imbalance. SamGoG constructs multiple GoGs through an efficient importance-based sampling mechanism and trains on them sequentially. This sampling mechanism incorporates the learnable pairwise similarity and adaptive GoG node degree to enhance edge homophily, thus improving downstream model quality. SamGoG can seamlessly integrate with various downstream GNNs, enabling their efficient adaptation for graph classification tasks. Extensive experiments on benchmark datasets demonstrate that SamGoG achieves state-of-the-art performance with up to a {\bf 15.66\%} accuracy improvement with {\bf 6.7$\times$} training acceleration. 
		\vspace{-0.5em}
	\end{abstract}
	
	\vspace{-0.7em}
\section{Introduction}
\vspace{-0.7em}
Graph classification is a supervised learning task that seeks to learn a predictive function $f: g \rightarrow \mathcal{C}$ on a set of graphs $\mathbb{G} = \{g_i\}_{i=1}^{N}$, utilizing a labeled subset $\mathbb{G}_{\mathrm{L}} = \{g_i\}_{i=1}^{N_\mathrm{L}} \subset \mathbb{G}$. Each graph $g_i\in \mathbb{G}_\mathrm{L}$ is represented as $g_i=(A_i,X_i,y_i)$, where $A_i$ is the corresponding adjacent matrix, $X_i$ is the node feature matrix, and $y_i \in \mathcal{C}$ is the class label, with $\mathcal{C}$ being the set of possible classes.
The goal is to learn a function $f$ that generalizes to predict labels for unlabeled graphs in $\mathbb{G}_{U} = \mathbb{G} \setminus \mathbb{G}_{L}$, by leveraging their structures and features.
This task has wide-ranging applications in areas such as social network analysis~\cite{10.1145/3308558.3313488} and bioinformatics~\cite{jha2022prediction,Prot2Text}, facilitating tasks like community detection, molecular property prediction, and protein function analysis.
To this end, GNNs~\cite{kipf2017semisupervised,GAT2018} have become the leading method due to their ability to jointly model graph structural information and node features.

Despite the success of GNNs, graph classification under imbalanced conditions remains a significant challenge, primarily due to two common issues: {\bf class imbalance}, where some classes contain far more training graphs than others~\cite{liu2024classimbalancedgraphlearningclass,pan2013graph} (see \cref{fig:clsimbdes2}), and {\bf graph size imbalance}, where input graphs vary substantially in size (i.e., number of nodes)~\cite{topoimb2022,soltgnn2022} (see \cref{fig:clsimbdes3}).
Class imbalance biases models toward majority classes, degrading the performance of minority ones, while graph size imbalance
impairs performance on small graphs and hinders generalization under distribution shifts.
Existing methods fall into data-centric (e.g., class resampling, synthetic augmentation)~\cite{chawla2002smote,drummond2003c4,wu2021graphmixupimprovingclassimbalancednode,imbgnn2024} and algorithmic-centric (e.g., subgraph matching, pattern extraction)~\cite{soltgnn2022,imgkb2023,topoimb2022} categories, but both face computational bottlenecks and limited effectiveness in handling imbalance.
Recent advances highlight the potential of the Graph-of-Graphs ({\bf GoG}) paradigm~\cite{g2gnn2022}, which represents each input graph as a GoG node in a higher-level graph, with edges between GoG nodes defined via pairwise similarity between corresponding input graphs. 
This inter-graph representation enables richer supervision: by embedding minority-class or small graphs within a broader context of similar graphs, GoG-based methods enhance representation learning beyond what is possible using only
 intra-graph features.

\begin{floatingfigure}[r]{0.5\textwidth}
	\vspace{-30pt}
	\begin{minipage}{0.5\textwidth}
	\subfigure[Class Imbalance \label{fig:clsimbdes2}]{
	\includegraphics[width=0.48\textwidth]{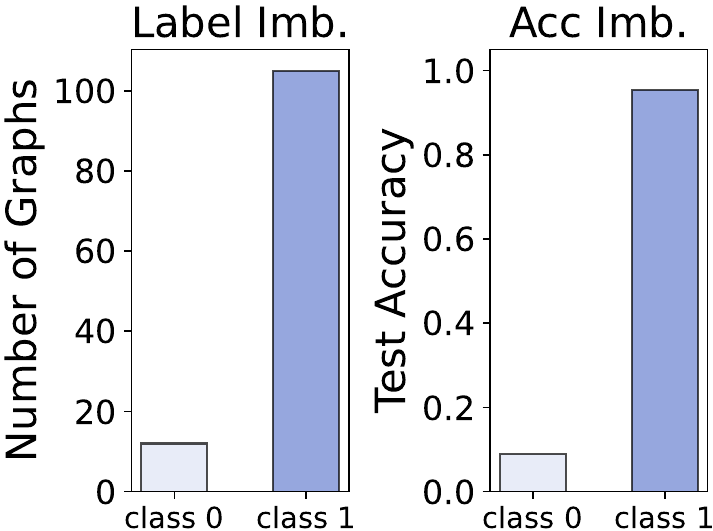}
	}%
	\subfigure[Graph Size Imbalance \label{fig:clsimbdes3}]{
	\includegraphics[width=0.48\textwidth]{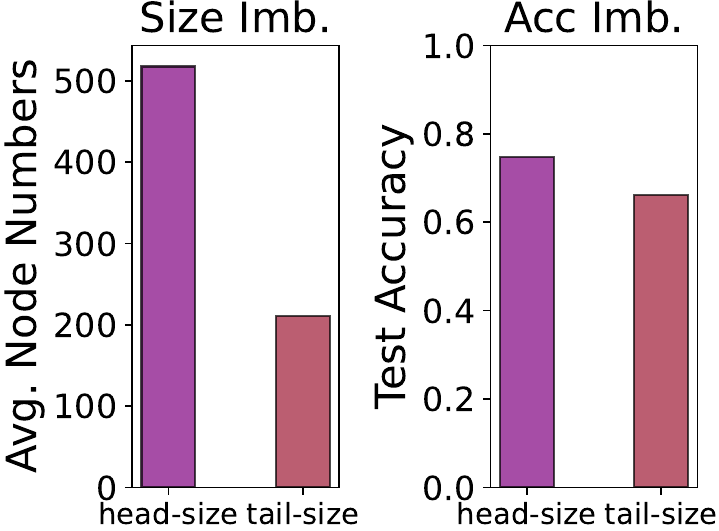}
	}
	\vspace{-1.0em}
	\caption{{\bf Imbalanced training data results in biased model performance in D\&D~\cite{wlstkernel} (GIN \cite{xu2018powerful})}: In (a), with only 10\% of training graphs in class 0, the model shows significantly lower accuracy in that class.
In (b), splitting test graphs by size shows that larger (head-size) graphs are generally easier to classify.
}

\vspace{-5pt}
		\label{fig:clsimbdes}
	\end{minipage}
\end{floatingfigure}

However, existing GoG approaches~\cite{g2gnn2022,imbgnn2024} encounter three major challenges that prevent them from fully harnessing the potential of GoG:
\emph{\bf Challenge 1:} Existing GoG methods treat edge homophily (see in~\cref{def:homophily}) purely as a post hoc evaluation metric rather than incorporating it as an optimization objective during construction so that the resulting GoGs lack theoretical guarantees and capability to capture inter-graph supervision.
\emph{\bf Challenge 2:} The construction of GoG edges relies heavily on pairwise graph similarity, 
which is typically computed using graph kernels (e.g., Weisfeiler-Lehman subtree~\cite{wlstkernel} and shortest path~\cite{spkernel}) that prioritize structural features over semantic ones.
{This struggles to precisely represent semantic relationships between graphs}
and incurs high computational cost (e.g., {\small$\mathcal{O}(N^2 \overline{|V|}^4)$}, where $N$ is the number of input graphs and {\small$\overline{|V|}$} is the average number of nodes per graph).
\emph{\bf Challenge 3:} 
Existing methods adopt a {one-shot $k$NN} strategy for GoG construction, assigning a fixed number of neighbors $k$ to each GoG node. However, this fixed $k$ makes the resulting GoG's edge homophily highly sensitive to the choice of $k$\footnote{When applying $\text{G}^2\text{GNN}$~\cite{g2gnn2022} with the shortest path kernel to the PTC-MR~\cite{ptcmr} and DHFR~\cite{sutherland2003spline} datasets (containing 344 and 756 graphs, respectively), increasing $k$ from 1 to 10 leads to a notable drop in edge homophily—by 9\% on PTC-MR and 15\% on DHFR.}.
Moreover, the one-shot construction introduces inductive biases and prevents the model from dynamically refining inter-graph relationships, underscoring the need for a more flexible and iterative GoG construction process.

To address these challenges, we propose {\bf SamGoG},  
{a homophily-driven framework, designed to efficiently construct multiple GoGs by importance-sampling in the presence of class and graph size imbalance and trains on them sequentially.}
{\bf For Challenge 1}, we formulate GoG construction as a constrained optimization problem where node degrees are adaptively assigned (\cref{sec:obj}). 
We prove that prioritizing neighbors with {better similarity quality} can increase expected edge homophily.
{\bf For Challenge 2}, we investigate a learnable pairwise similarity function that jointly encodes structural and semantic features. This not only eliminates the overhead of kernel-based methods but also improves edge homophily by guiding the sampler to favor same-class neighbors.
{\bf For Challenge 3}, we propose: a) an adaptive GoG node degree allocation strategy that adjusts neighborhood size for each GoG node to achieve higher homophily under class and graph size imbalance (\cref{sec:obj}); and b) a dynamic and parallel sampling method that generates multiple GoGs {to capture a wider range of graph-level interactions} during training. This replaces one-shot $k$NN construction, reduces inductive bias, and allows inter-graph supervision (\cref{sec:sampling}) to evolve throughout training.

Our contributions can be summarized as follows.
{\bf First}, we propose SamGoG, the first framework to formulate GoG construction as a constrained homophily maximization problem, theoretically establishing the relationship between node degree allocation and homophily maximization.
Our analysis demonstrates that degree distributions prioritizing nodes with better quality of similarity enhance homophily stability under~\cref{thm:score2k}, providing a theoretical foundation for topology-adaptive sampling aligned with {the class and graph size imbalances}.
{\bf Second}, we design three complementary mechanisms for GoG construction: 1) a learnable pairwise graph similarity function replacing graph kernels, which serves as the importance metric for sampling, 2) an adaptive GoG node degree allocation strategy replacing fixed node degree in the $k$NN method, and 3) a dynamic sampling method replacing one-shot $k$NN construction to generate multiple GoGs during training. {Compared to baselines, ours improves edge homophily by over 5\%, while achieving a 4-6 order of magnitude reduction in similarity computation time.}
{\bf Third}, SamGoG achieves state-of-the-art (SOTA) performance on imbalanced graph classification benchmarks~\cite{iglbench}, demonstrating significant improvements across diverse datasets. Specifically, it outperforms baselines with an average accuracy increase of {5.07\%} in 24 class imbalance experiments across 4 datasets, and an average accuracy gain of {4.13\%} in 24 graph size imbalance experiments on the same datasets. {Moreover, SamGoG achieves up to a 6.7$\times$ acceleration over baselines}. These results underscore SamGoG’s outstanding performance in both accuracy and efficiency.
\vspace{10pt}
	\vspace{-0.7em}
\section{Preliminary}
\vspace{-0.7em}
\textbf{Graph-of-Graphs.}
The GoG represents a set of input graphs as nodes of a higher-order graph, therefore converting the problem of {\it graph classification} (assigning labels to input graphs) into a {\it semi-supervised node classification} task (assigning labels to GoG nodes). 

{Formally, given a set of $N$ input graphs $\mathbb{G} = \{g_i\}_{i=1}^{N}$, let $\mathbb{G}_\mathrm{L}$ and $\mathbb{G}_\mathrm{U} = \mathbb{G} \setminus \mathbb{G}_{L}$ denote the labeled and unlabeled subsets, respectively.
Each input graph $g_i=(A_i,X_i,y_i) \in \mathbb{G}$ has $s_i$ nodes (i.e., as graph size). The GoG is constructed as a high-order graph $G = (V, E, H, Y)$, where: (1) each GoG node $v_i \in V$ corresponds to an input graph $g_i \in \mathbb{G}$; (2) the  GoG node feature $h_i \in H$ is encoded as $h_i = \operatorname{Readout}\left(\mathcal{M}(A_i, X_i)\right)$, where $\mathcal{M}(\cdot)$ denotes a GNN for node encoding and $\operatorname{Readout}(\cdot)$ (e.g., mean/sum pooling~\cite{10.5555/2969442.2969488,xu2018powerful}) aggregates node embeddings;
(3) GoG edges $E$ capture relationships between input graphs (e.g., similarity);
and (4) labels of GoG nodes $Y = \{y_i\}_{i=1}^N$ correspond directly to the labels of the original input graphs.
Only the labels associated with the labeled subset $\{y_j \,|\, g_j \in \mathbb{G}_L\}$ are observable, while the labels $\{y_k \,|\, g_k \in \mathbb{G}_U\}$ remain unobserved and are the targets to be predicted.
}

The core of GoG construction lies in building the edge set $E$ based on a precomputed pairwise graph similarity matrix $S \in \mathbb{R}^{N \times N}$. This process involves two key steps: 1) {\bf Neighborhood Selection}: determine the count of edge connections $k_i$ for each GoG node $v_i$, defining its neighborhood size.
2) {\bf Edge Materialization}: for each $v_i$, update edge set $E$ by adding directed edges from $v_i$ to its $k_i$ most similar counterparts in the GoG, as determined by the pairwise similarity matrix $S$.

\textbf{Related Works.} GoG extends the conventional GNN paradigm
~\cite{10.1145/2623330.2623643}. Early works like NoN~\cite{gu2021modelingmultiscaledatanetwork} and GoGNN~\cite{Wang_2020} introduce GoG for graph classification and link prediction, respectively. G\textsuperscript{2}GNN~\cite{g2gnn2022} advances this idea by constructing a $k$NN-based GoG using graph similarity, alleviating class imbalance in graph classification via inter-graph message passing.
Building on G\textsuperscript{2}GNN, ImbGNN~\cite{imbgnn2024} introduces a similarity-aware graph random walk to extract core subgraphs, enhancing the reliability of kernel-based similarity under graph size imbalance.
This enhancement addresses G\textsuperscript{2}GNN's limitation in underestimating similarity between graphs of different sizes but results in poorer efficiency due to the cost of extracting core subgraphs and more computation in pairwise similarity.

	\vspace{-0.7em}
\section{SamGoG}
\vspace{-0.7em}

\begin{floatingfigure}[r]{0.6\linewidth}
	\vspace{-20pt}
	\begin{minipage}{0.6\linewidth}
		\begin{algorithm}[H]
			\caption{SamGoG Pipeline}
			\label{alg:training}
			\begin{algorithmic}[1]
				\STATE Pre-compute GoG node degree $k_i$ for all training graphs via three rules
				\FOR{each training epoch}
				\STATE Obtain GoG node representations $H$
				\STATE Calculate pairwise similarity matrix $S$ based on $H$ via~\cref{eq:similarity}
				\STATE Perform parallel sampling and construct GoGs based on node degrees $\{k_i\}$ and the similarity matrix $S$
				\STATE Obtain label prediction based on downstream models
				\STATE Calculate loss function and back propagation
				\ENDFOR
			\end{algorithmic}
		\end{algorithm}
	\end{minipage}
\end{floatingfigure}
{\bf Framework.} This section introduces the SamGoG, a sampling-based GoG framework for imbalanced graph classification by optimizing edge homophily, as shown in \cref{fig:overview}.
The detailed process is depicted in~\cref{alg:training}.
Initially, SamGoG calculates the GoG node degree ($k_i$) for all training graphs based on the three rules investigated in~\cref{sec:degree}.
Then, SamGoG embeds the input graphs with GNN encoders to obtain graph representations, i.e., GoG node embeddings (Line 3).
Next, a multi-layer perceptron computes pairwise graph similarity matrix $S \in \mathbb{R}^{N \times N}$ between graph representations (Line 4).
We then perform parallel sampling from an implicit complete graph\footnote{The implicit complete graph, where edge weights are defined by pairwise similarity.~\cite{g2gnn2022,imbgnn2024,simba}.}. guided by the node degrees $\{k_i\}$ and the similarity matrix $S$ (Line 5).
Each GoG node's degree determines how many neighbors it selects in the sampled sub-GoG, while $S$ defines the connection probabilities between node pairs.
This process generates multiple sub-GoGs.
Finally, these sampled sub-GoGs are used for training downstream backbone models (Line 6).

{The rest of this section proceeds as follows. {\S\ref{sec:obj}} formulates the GoG construction optimization objective based on edge homophily. {\S\ref{sec:keyc}} presents the methods of pairwise graph similarity computation
and GoG node degree allocation.
{\S\ref{sec:sampling}} analyzes subgraph training effectiveness on sampled GoGs and the complexity of several key components in the framework.

\begin{figure*}[t]
	\vspace{-0.5em}
	\centering
	\includegraphics[width=\textwidth]{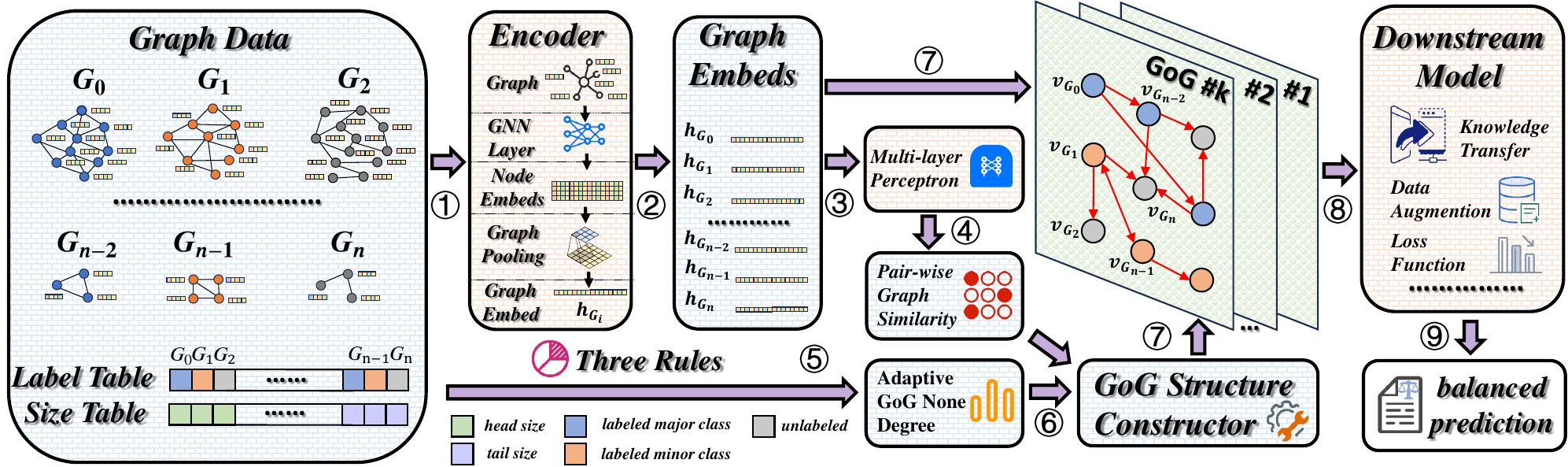}
	\vspace{-0.5em}
	\caption{{\bf Overview of SamGoG}:
		{\bf \ding{172}} Input graphs to elemental GNN encoders;
		{\bf \ding{173}} Output graph representations;
		{\bf \ding{174}} Use an MLP to compute the classification logits;
		{\bf \ding{175}} Compute pair-wise graph similarity matrix;
		{\bf \ding{176}} Adaptively allocate node degree for GoG nodes by class and size information;
		{\bf \ding{177}} Construct GoGs by importance sampling based on pair-wise graph similarity and adaptive node degree;
		{\bf \ding{178}} Combine GoG node features and sampled structure;
		{\bf \ding{179}} Train node classification model on constructed GoGs;
		{\bf \ding{180}} Output final result.}
	\vspace{-2.0em}
	\label{fig:overview}
\end{figure*}

\vspace{-0.7em}
\subsection{Graph-of-Graphs Construction Objective}
\label{sec:obj}
\vspace{-0.7em}

We begin by defining the objective of GoG construction.
Edge homophily is a widely acknowledged metric for evaluating the quality of GoG construction in existing works~\cite{g2gnn2022,imbgnn2024}.
Motivated by this, our goal is to maximize {\bf edge homophily}~\cite{zhu2020beyond} of constructed GoGs.
It is worth noting that homophily can be defined in multiple ways~\cite{pei2020geomgcngeometricgraphconvolutional,zhu2020beyond,lim2021newbenchmarkslearningnonhomophilous}. To avoid ambiguity, we use ``\textbf{homophily}'' to specifically refer to edge homophily throughout this paper. Homophily is defined as follows:

\begin{definition}\label{def:homophily}
(\textbf{Homophily~\cite{zhu2020beyond}}): Given a GoG graph $G = (V,E,H,Y)$, the homophily of $G$ ($\text{homo}(G)$)is defined as the fraction of edges linking two GoG nodes with the same label. Formally, we have homophily:
	\begin{equation}
		\label{equ:ehomo0}
		\mathrm{homo}(G) = \frac{1}{|E|} \sum_{(v_i, v_j) \in E} \mathbb{I}(y_i = y_j)
	\end{equation}
	, where $|E|$ denotes the number of edges. $\mathbb{I}(\cdot)$ denotes the indicator function that returns 1 when $y_i = y_j$, and 0 otherwise.
\end{definition}

However, directly optimizing the homophily defined in Definition~\ref{def:homophily} is an optimization problem without constraints on node degrees, {which harms GNN performance by incurring message-passing bottlenecks}~\cite{10.1145/3340531.3411910,tailgnn}.
In response, we design a constrained optimization that maximizes homophily {subject to node degree constraints}, as shown below.

\vspace{-1.0em}
\begin{equation}
	\label{equ:ehomo}
	\begin{aligned}
		\text{maximize} & \quad \mathbb{E}(\text{homo}(G)) = \frac{\sum_i{k_i \cdot \text{prob}(v_i,y_i)}}{\sum_i k_i} \\
		\text{s.t.} & \quad k_{\text{min}} \leq k_i \leq k_{\text{max}} \ \forall i \quad \text{and} \quad \sum_i k_i = N\bar{d}
	\end{aligned}
\end{equation}
\vspace{-1.0em}

, where $\text{prob}(v_i, y_i) = \sum_{y_j \in Y} \mathbb{I}(y_j = y_i) \cdot \frac{S[i,j]}{\sum_k S[i,k]}$ denotes the homophily probability that GoG node $v_i$ connects to neighbors with the same label $y_i$, based on the similarity matrix $S$.
Here, $\bar{d}$ is the target average GoG node degree, which is a pre-defined parameter. $k_{\text{min}}$ and $k_{\text{max}}$ are user-defined bounds that constrain the range of GoG node degrees. $N$ is the number of input graphs, which is also the number of GoG nodes.
$\sum_i k_i = N\bar{d}$ ensures that the sum of GoG node degrees is a fixed value, i.e., $N\bar{d}$.

{\bf Key Insights.} Based on this optimization objective in~\cref{equ:ehomo}, two critical factors emerge as essential for improving the quality of sampling-based GoG construction.
{\bf First}, improving the quality of pairwise graph similarity, i.e.,  enhancing the similarity among GoG nodes sharing the same label, contributes to a higher value of $\text{prob}(v_i,y_i)$.
{\bf Second}, optimizing the distribution of GoG node degrees $\{k_i\}$ can further improve homophily: GoG nodes with higher $\text{prob}(v_i,y_i)$ should be assigned higher degrees $k_i$ to maximize the overall homophily.

This leads us to a fundamental partial ordering principle, which provides a simplified yet effective solution to the optimization problem:

\begin{lemma}
	For any two GoG nodes $v_i$ and $v_j$ with degrees $k_i$ and $k_j$, if $\mathbb{E}(\mathrm{prob}(v_i,y_i)) > \mathbb{E}(\mathrm{prob}(v_j,y_j))$, then achieving optimal homophily in~\cref{equ:ehomo} requires $k_i > k_j$. Conversely, if $\mathbb{E}(\mathrm{prob}(v_i,y_i)) \leq \mathbb{E}(\mathrm{prob}(v_j,y_j))$, then $k_i \leq k_j$ leads to better homophily.
	\label{thm:score2k}
\end{lemma}
	Lemma \ref{thm:score2k} is proven in \cref{sec:appproofhomo}.
\cref{thm:score2k} establishes a relationship between homophily, degree allocation, and expected homophily probability. This principle allows us to prioritize higher degrees to GoG nodes with higher expected $\mathrm{prob}(v_i,y_i)$, thereby improving the overall homophily in the GoG construction process.

	\vspace{-0.7em}
\subsection{Key Components}
\label{sec:keyc}
\vspace{-0.5em}
{\subsubsection*{Pairwise Graphs-of-Graph Node Similarity Calculation}}
\vspace{-0.5em}
\label{sec:simi}
As shown in Algorithm~\ref{alg:training}, we first extract the feature repsentations of GoG nodes by employing a GNN (e.g., GIN~\cite{xu2018powerful} and GCN~\cite{kipf2017semisupervised}).
Through the message passing paradigm of GNNs, these feature representations capture the structural information present in the training graphs.
Based on these representations, we further use an MLP-based method to encode the GoG nodes and get initial classification logits ($\mathrm{logits}(g_i)$). To further take into account the label information of the training graphs, we calculate the similarity by incorporating labeled certainty and unlabeled uncertainty, as shown in \cref{eq:similarity}.
\begin{equation}
		\label{eq:similarity}
\begin{aligned}
	S[i, j] &= P(\hat{y}_i) \cdot P(\hat{y}_j)
	= \sum_{c \in \mathcal{C}} P(\hat{y}_i = c)\, P(\hat{y}_j = c) \\
	P(\hat{y}_i) &=
	\begin{cases}
		\text{onehot}(y_i), & g_i \in \mathbb{G}_{\mathrm{L}} \\
		\text{Softmax}(\text{logits}(g_i)), & g_i \notin \mathbb{G}_{\mathrm{L}}
	\end{cases}
\end{aligned}
\end{equation}
where $\mathrm{logits}(g_i) \in \mathbb{R}^{\mid\mathcal{C}\mid}$, $P(\hat{y}_i) \in \mathbb{R}^{\mid\mathcal{C}\mid}$, and $\sum_{c \in \mathcal{C}} P(\hat{y}_i=c) = 1$.
\cref{eq:similarity} adjusts the GoG node similarity by incorporating labeled certainty and unlabeled uncertainty. For labeled graphs ($\{g_i | g_i \in \mathbb{G}_{\mathrm{L}}\}$), we preserve ground-truth certainty through one-hot encoding, while for unlabeled graphs ($\{g_i | g_i \notin \mathbb{G}_{\mathrm{L}}\}$), we maintain probability distributions for the prediction label $\hat{y}_i$. For two graphs $g_i$ and $g_j$, the joint probability of sharing class $c$ is $P(\hat{y}_i=c,\, \hat{y}_j=c) = P(\hat{y}_i=c)P(\hat{y}_j=c)$ because of the independent and identically distributed (i.i.d.) assumption as shown in~\cite{elkan2001foundations,ling2008cost,zhou2006training}.
The full similarity matrix $S \in \mathbb{R}^{N \times N}$ can be computed efficiently as a parallel matrix operation ($S = PP^T$).
Then, $S$ can be further used to guide importance sampling for GoG construction as shown in Line 5 of Algorithm~\ref{alg:training}.

\vspace{-0.5em}
{\subsubsection*{Graphs-of-Graph Node Degree Pre-allocation}}
\label{sec:degree}
\vspace{-0.5em}
\cref{thm:score2k} demonstrates that enhancing homophily via degree allocation leads to improved model accuracy.
Inspired by this, we propose three rules to allocate GoG node degrees based on these distinct perspectives: the labeled and unlabeled parts: the majority and minority class in the labeled part; head-size and tail-size graphs of the unlabeled part.
Based on the constraints in \cref{equ:ehomo}, we first assign $k_\mathrm{min}$ degrees to each GoG node. The remaining $\Delta = N(\bar{d} - k_\mathrm{min})$ degrees are then allocated according to the following rules.

{\textbf{Motivation 1: Labeled and Unlabeled GoG Nodes.}
		First, we design the rule based on the differences between labeled and unlabeled GoG nodes.
	Since labeled nodes incorporate accurate class information during similarity computation, they tend to achieve better representation than their unlabeled counterparts.
	Motivated by this, we design the label priority rule as follows.
	
	\begin{myrule}[\bf Label Priority]
		\label{rule:rule1}
		Given a hyperparameter ratio $\rho_1$ ($\rho_1 > 1$), we divide $\Delta$ degrees to $\Delta_\mathrm{L}$ and $\Delta_\mathrm{U}$ between all the labeled and unlabeled nodes that $\frac{\Delta_\mathrm{L} / N_\mathrm{L}}{\Delta_\mathrm{U} / N_\mathrm{U}} = \rho_1$. {$N_\mathrm{L}$ and $N_\mathrm{U}$ are labeled graph numbers and unlabeled graph numbers, respectively.} 
	\end{myrule}
	We can explain why Rule 1 works from the perspective \cref{thm:score2k}.
	Because our focus is solely on the impact of labeling, we control for other variables by considering a pair of GoG nodes, $v_i$ (labeled) and $v_j$ (unlabeled), that share the same true label. 
Under this setup, we find that $\mathbb{E}(\mathrm{prob}(v_i,y_i)) > \mathbb{E}(\mathrm{prob}(v_j,y_j))$ (Please refer to Appendix~\ref{proof:r1} for details). Therefore, according to Lemma~\ref{thm:score2k}, assigning a higher degree to the labeled GoG node (i.e., setting $k_i > k_j$) will lead to increased homophily.

	\textbf{Motivation 2: Major and Minor Classes in the Labeled GoG Nodes.} Secondly, we design the rule based on the differences between major and minor classes in the labeled part. We assume that there is no shift in the class distribution between the labeled and unlabeled portions. As shown in Figure~\ref{fig:clsimbdes}(a), major classes achieve higher accuracy compared to minor classes, which shows that major classes have better representation. Motivated by this, we design the class imbalance rule as follows.

	\begin{myrule}[\bf Class Imbalance]
		\label{rule:rule2}
		Given a hyperparameter ratio $\rho_2 > 1$, we divide $\Delta_\mathrm{L}$ degrees to $\Delta_\mathrm{major}$ and $\Delta_\mathrm{minor}$ between all labeled nodes in majority and minority classes, such that $\frac{\Delta_\mathrm{major}}{\Delta_\mathrm{minor}} = \rho_2$.
	\end{myrule}
Similarly, we can also explain why Rule 2 is effective through the lens of \cref{thm:score2k}. 
Consider two labeled GoG nodes, $v_i$ with label $y_i$ from the majority class and $v_j$ with label $y_j$ from the minority class.
We can get $\mathbb{E}(prob(v_i,y_i)) > \mathbb{E}(prob(v_j,y_j))$ as detailed in Appendix~\ref{proof:r2}. According to~\cref{thm:score2k}, assigning a higher degree to the GoG node from the majority class (i.e., setting $k_i > k_j$) will improve overall homophily.

{\textbf{Motivation 3: Size Distributions in the Unlabeled GoG Nodes.}
	Thirdly, we design the rule based on the differences in the size distributions of the unlabeled GoG nodes.
	
	As shown in Figure~\ref{fig:clsimbdes}(b), The imbalanced graph size leads to imbalanced classification performance.
	GNNs struggle to generalize across graph sizes not well represented in the training set, particularly when extrapolating from small to big graphs.
In scenarios without a significant size distribution shift, head-size graphs typically achieve better performance. 
ImbGNN~\cite{imbgnn2024} highlights that for graphs with significant size differences but similar semantic information and the same true label, graph kernel functions tend to underestimate their similarity, hindering cross-size knowledge transfer in GoG-based methods. 
{To address this, we introduce a window-based strategy, in which higher degrees are assigned to GoG nodes whose sizes better align with the training graph size distribution. The measure of alignment in size is quantified using a size window.
	
}

		\begin{myrule}[\bf Size Adaptation]
		\label{rule:rule3}
Let $w_\mathrm{train}=\{s_k \big| g_k \in \mathbb{G}_{L}\}$ denote the size distribution of training graphs and $2r$ be the window width. 
		For each unlabeled node $v_i$ with size $s_i$, define its size window $w_i = [s_i - r, s_i + r]$. The allocated degrees $\Delta_\mathrm{U}$ are distributed in proportion with $|w_\mathrm{train} \cap w_i|$ as weight, where $|w_\mathrm{train} \cap w_i|$ counts training graphs within $v_i$'s size window.
	\end{myrule}
	
	This rule is based on \cref{thm:score2k}. For two unlabeled GoG nodes $v_i$ and $v_j$, if $|w_\mathrm{train} \cap w_i| > |w_\mathrm{train} \cap w_j|$, then $\mathbb{E}(\mathrm{prob}(v_i,y_i)) > \mathbb{E}(\mathrm{prob}(v_j,y_j))$ (\cref{proof:r3}). 
According to Lemma~\ref{thm:score2k}, assigning a higher degree to $v_i$ (i.e., $k_i > k_j$) improves homophily by prioritizing nodes whose sizes are better aligned with the training distribution.
	
	Since the entire allocation process does not require the learned node representations, the GoG node degree allocation can be performed offline, as shown in Line 1 of Algorithm~\ref{alg:training}.
	
	In summary, the procedure is as follows: (1) allocate degrees between labeled and unlabeled nodes (Rule 1); (2) adjust class-wise degrees in the labeled subset (Rule 2); and (3) apply Rule 3 to the unlabeled subset.
	Finally, the allocated degrees ($\{k_i\}$) are further used to guide importance sampling for GoG construction as shown in Line 5 of Algorithm~\ref{alg:training}.
	
	}

	\vspace{-0.7em}
\subsection{Analysis}\label{sec:sampling}
\vspace{-0.7em}
{\bf Sampling Analysis.}
To establish theoretical guarantees, we analyze the relationship between {sample size $t$}
 (number of GoG constructions) and downstream performance.

\begin{theorem}\label{thm:samplingcon}
	\textbf{Convergence of Sampling}. Every sampled subgraph-based gradient constitutes an unbiased estimator of the full-graph objective's gradient. Therefore, GCN parameters $\Theta^{(t)}$ converge in expectation to $\Theta_{\mathrm{full}}$, satisfying $\lim_{t \to \infty} \mathbb{E}[H^{(t)}] = H_{\mathrm{full}}$, where $H_{\mathrm{full}}$ denotes node representations trained on the complete similarity-weighted graph.
\end{theorem}
Theorem \ref{thm:samplingcon} is proven in~\cref{proof:sampling}.
Theorem \ref{thm:samplingcon} shows that the performance of GoG sampling-based method depends on the number of GoG constructions.
When $t$ is sufficiently large, our sampling-based method can closely approximate full-graph training, which leverages all available information.
However, training on the complete graph, while allowing for greater information utilization, also results in low training efficiency.
In contrast, we achieve better efficiency by optimizing importance-sampling with adaptive GoG node degree and training in parallel. 
Therefore, the sampling-based method is a better choice.

\begin{theorem}\label{thm:samplingvr}
	\textbf{Variance Analysis with $t$.} With diminishing learning rates, \(\mathrm{Var}(\Theta^{(t)})\) remains bounded with scaling behavior \(\mathcal{O}(1/t)\). Consequently, $\mathrm{Var}\bigl(H^{(t)}\bigr) = \mathcal{O}\!\Bigl(\frac{1}{t}\Bigr)$, causing final embeddings to concentrate around the full-graph solution as \( t \) increases.
\end{theorem}
	Theorem \ref{thm:samplingvr} is proven in~\cref{proof:sampling}.
We additionally analyze the sampling based on variance as shown in Theorem \ref{thm:samplingvr}. This variance-reduction property demonstrates that increasing {sample size $t$}
 reduces inductive bias, consistent with the law of large numbers. 
Our sampling framework is also highly scalable in practice with GPU-optimized operations. It completes GoG construction for batches of over 1,000 graphs within milliseconds, which is several orders of magnitude faster than computing individual graph embeddings.

{\bf Complexity Analysis.}
{
	 In SamGoG, we adopt TailGNN~\cite{tailgnn} as the downstream model for its effectiveness in knowledge transfer from high-degree to low-degree nodes.} The additional computational overhead primarily stems from three components {in GoG construction}: pairwise graph similarity calculation, adaptive GoG node degree allocation, and the sampling-based construction of the GoG structure. Here, $N$ represents the number of graphs, and $|\mathcal{C}|$ denotes the number of classes.
For similarity calculation, we compute similarity via matrix multiplication, with a time complexity of $\mathcal{O}(N^2|\mathcal{C}|)$, which is significantly lower than graph kernel methods. The GoG node degree allocation algorithm assigns the total number of edges based on predefined rules, resulting in a time complexity of $\mathcal{O}(N)$. For sampling-based construction of the GoG structure, its time complexity on a single GoG node is $\mathcal{O}(N)$, primarily due to normalization and cumulative distribution function calculations. Since neighborhoods of all $N$ GoG nodes are sampled and aggregated in parallel, the overall time complexity for constructing the GoG remains $\mathcal{O}(N)$. 
In comparison, the overall time complexity of SamGoG is $\mathcal{O}(N^2|\mathcal{C}|)$, while that of existing GoG methods with kernel-based similarity~\cite{g2gnn2022,imbgnn2024} is $\mathcal{O}(N^2 |\mathcal{V}|^3)$~\cite{g2gnn2022}, and other baseline methods~\cite{datadec2023,soltgnn2022,topoimb2022,imgkb2023} are $\mathcal{O}(\sum_{i=1}^N(|\mathcal{V}_i| + |\mathcal{E}_i|))$, where $|\mathcal{V}_i|$ and $|\mathcal{E}_i|$ represent the number of nodes and edges in each graph.
	\vspace{-0.7em}
\section{Experiments}
\vspace{-0.7em}
In this section, we conduct extensive experiments to answer the following research questions: \textbf{(RQ1)} Can SamGoG effectively enhance the overall performance of imbalanced graph classification? \textbf{(RQ2)} Can SamGoG help improve the classification performance on minority class and tail-size graphs? \textbf{(RQ3)}
Can SamGoG construct GoGs with ideal edge homophily? \textbf{(RQ4)} Is SamGoG efficient enough on large-scale graph data?
\vspace{-0.7em}
\subsection{Experiment Setup}
\label{sec:expsetup}
\vspace{-0.7em}
\textbf{Benchmark Settings.} We evaluate SamGoG on standard imbalanced classification benchmarks, IGL-Bench~\cite{iglbench}. Following IGL-Bench, we adopt TUDatasets~\cite{tudatasets} and ogbg-molhiv~\cite{ogbbenchmark} as datasets; details are provided in Appendix~\ref{app:datasets}. We use the predefined splits of each dataset with varying class/graph size imbalance ratios (defined in~\cref{eq:ratioclass,eq:ratiosize} in Appendix~\ref{sec:imbdes}). Our implementation uses either GIN~\cite{xu2018powerful} or GCN~\cite{kipf2017semisupervised} as base encoders and employs TailGNN~\cite{tailgnn} as the downstream model. Other implementation details are provided in Appendix~\ref{sec:experiment}.

\textbf{Baselines.} We compare against all graph classification baselines in IGL-Bench~\cite{iglbench}, with additional comparisons of our GoG construction against kernel-based methods~\cite{spkernel,wlstkernel,randomwalkkernel,gsamk} in $\text{G}^2\text{GNN}$~\cite{g2gnn2022}. 

\begin{table}[!ht]
\vspace{-10pt}
\centering
\caption{Accuracy (\% ± stddev) on datasets with changing class imbalance levels over 10 runs, encoded by GIN / GCN. {\colorbox{firstplace}{\color{red}\bf Red}} and  {\colorbox{secondplace}{Blue}} highlight the best and second-best results respectively.}
\label{tab:main_graph_cls_acc_all}
\begin{minipage}[t]{0.49\linewidth}
\centering
\resizebox{\textwidth}{!}{  
	\setlength{\tabcolsep}{0.8em}
	{\footnotesize
		\begin{tabular}{lcccc}
			\toprule
			\textbf{Algorithm} & \textbf{PTC-MR}   & \textbf{D\&D} & \textbf{IMDB-B} & \textbf{REDDIT-B} \\
			\midrule
			\multicolumn{5}{c}{Balance (${\rho_{\text{class}}}=5:5$)}\\
			\midrule
			GIN (bb.)~\cite{xu2018powerful} & 50.43\scalebox{0.75}{±2.69} & 64.04\scalebox{0.75}{±3.79} & 66.05\scalebox{0.75}{±2.57} & 76.66\scalebox{0.75}{±4.80} \\
			G\textsuperscript{2}GNN~\cite{g2gnn2022} & 53.70\scalebox{0.75}{±3.87}  & 66.07\scalebox{0.75}{±2.27} & 61.91\scalebox{0.75}{±3.77} & 72.34\scalebox{0.75}{±2.76} \\
			TopoImb~\cite{topoimb2022} & 50.91\scalebox{0.75}{±2.18} & \cellcolor{secondplace}{66.57\scalebox{0.75}{±3.81}} & \cellcolor{secondplace}{66.28\scalebox{0.75}{±1.85}} & 73.99\scalebox{0.75}{±1.18} \\
			DataDec~\cite{datadec2023} & \cellcolor{secondplace}{54.05\scalebox{0.75}{±4.85}} & 64.46\scalebox{0.75}{±1.88} & 64.09\scalebox{0.75}{±5.75} & \cellcolor{firstplace}{\bf\color{red}79.29\scalebox{0.75}{±8.18}} \\
			ImGKB~\cite{imgkb2023} & 53.48\scalebox{0.75}{±3.50} & 65.45\scalebox{0.75}{±2.88} & 50.16\scalebox{0.75}{±0.34} & 50.24\scalebox{0.75}{±0.29} \\
			SamGoG & \cellcolor{firstplace}{\bf\color{red}59.71\scalebox{0.75}{±2.12}} & \cellcolor{firstplace}{\bf\color{red}72.59\scalebox{0.75}{±1.93}} & \cellcolor{firstplace}{\bf\color{red}69.27\scalebox{0.75}{±1.53}} & \cellcolor{secondplace}{77.74\scalebox{0.75}{±0.50}} \\
			\midrule    
			\multicolumn{5}{c}{Low Imbalance ($\rho_{\text{class}}=7:3$)}\\
			\midrule
			GIN (bb.)~\cite{xu2018powerful} & 47.83\scalebox{0.75}{±2.95} & 51.05\scalebox{0.75}{±5.07} & 62.31\scalebox{0.75}{±3.99} & 61.10\scalebox{0.75}{±4.86} \\
			G\textsuperscript{2}GNN~\cite{g2gnn2022} & 51.88\scalebox{0.75}{±6.23}  & 56.29\scalebox{0.75}{±7.30} & 63.87\scalebox{0.75}{±4.64} & 69.58\scalebox{0.75}{±3.59}  \\
			TopoImb~\cite{topoimb2022} & 44.86\scalebox{0.75}{±3.52} & 49.97\scalebox{0.75}{±7.24} & 59.95\scalebox{0.75}{±5.19} & 59.67\scalebox{0.75}{±7.30} \\
			DataDec~\cite{datadec2023} & \cellcolor{secondplace}{55.72\scalebox{0.75}{±2.88}}  & 63.51\scalebox{0.75}{±1.62} & \cellcolor{secondplace}{67.92\scalebox{0.75}{±3.37}} & \cellcolor{secondplace}{78.39\scalebox{0.75}{±5.01}}  \\
			ImGKB~\cite{imgkb2023} & 50.11\scalebox{0.75}{±5.95}  & \cellcolor{secondplace}{65.85\scalebox{0.75}{±3.70}} & 47.74\scalebox{0.75}{±0.29} & 48.57\scalebox{0.75}{±2.14} \\
			SamGoG & \cellcolor{firstplace}{\bf\color{red}57.25\scalebox{0.75}{±2.85}} & \cellcolor{firstplace}{\bf\color{red}73.31\scalebox{0.75}{±1.99}} & \cellcolor{firstplace}{\bf\color{red}67.96\scalebox{0.75}{±0.57}} & \cellcolor{firstplace}{\bf\color{red}78.67\scalebox{0.75}{±1.44}} \\
			\midrule
			\multicolumn{5}{c}{High Imbalance ($\rho_{\text{class}}=9:1$)}\\
			\midrule
			GIN (bb.)~\cite{xu2018powerful} & 39.42\scalebox{0.75}{±1.87}  & 41.54\scalebox{0.75}{±6.57} & 53.57\scalebox{0.75}{±3.21} & 55.56\scalebox{0.75}{±7.85} \\
			G\textsuperscript{2}GNN~\cite{g2gnn2022} & 46.52\scalebox{0.75}{±9.94} & 55.38\scalebox{0.75}{±15.60} & 59.44\scalebox{0.75}{±6.49} & 63.22\scalebox{0.75}{±4.67} \\
			TopoImb~\cite{topoimb2022} & 39.42\scalebox{0.75}{±1.24} & 39.12\scalebox{0.75}{±1.62} & 47.75\scalebox{0.75}{±3.73} & 51.58\scalebox{0.75}{±4.69} \\
			DataDec~\cite{datadec2023} & \cellcolor{secondplace}{58.69\scalebox{0.75}{±3.10}} & \cellcolor{secondplace}{65.77\scalebox{0.75}{±2.71}} & \cellcolor{secondplace}{66.30\scalebox{0.75}{±6.70}} & \cellcolor{secondplace}{77.72\scalebox{0.75}{±5.12}} \\
			ImGKB~\cite{imgkb2023} & 44.24\scalebox{0.75}{±5.65} & 59.99\scalebox{0.75}{±7.57} & 47.08\scalebox{0.75}{±3.72} & 51.25\scalebox{0.75}{±5.10} \\
			SamGoG & \cellcolor{firstplace}{\bf\color{red}58.81\scalebox{0.75}{±2.30}} & \cellcolor{firstplace}{\bf\color{red}71.94\scalebox{0.75}{±1.73}} & \cellcolor{firstplace}{\bf\color{red}66.66\scalebox{0.75}{±2.25}} & \cellcolor{firstplace}{\bf\color{red}78.41\scalebox{0.75}{±1.63}} \\
			\bottomrule
		\end{tabular}%
	}
}
\label{tab:main_graph_cls_gin_acc_all}
\end{minipage}
\hfill  
\begin{minipage}[t]{0.49\linewidth} 
\centering
\resizebox{\textwidth}{!}{  
	\setlength{\tabcolsep}{0.8em}
	{\footnotesize
		\begin{tabular}{lcccc}
			\toprule
			\textbf{Algorithm} & \textbf{PTC-MR}   & \textbf{D\&D} & \textbf{IMDB-B} & \textbf{REDDIT-B} \\
			\midrule
			\multicolumn{5}{c}{Balance (${\rho_{\text{class}}}=5:5$)}\\
			\midrule
			GCN (bb.)~\cite{kipf2017semisupervised} & 48.62\scalebox{0.75}{±7.12}  & 63.97\scalebox{0.75}{±4.09} & \cellcolor{secondplace}{59.76\scalebox{0.75}{±1.78}} & 66.65\scalebox{0.75}{±1.06} \\
			G\textsuperscript{2}GNN~\cite{g2gnn2022} & 44.35\scalebox{0.75}{±4.32} & 61.47\scalebox{0.75}{±7.70} & 58.35\scalebox{0.75}{±3.47} & 67.02\scalebox{0.75}{±2.55} \\
			TopoImb~\cite{topoimb2022} & 50.61\scalebox{0.75}{±6.39} & 56.13\scalebox{0.75}{±4.61} & 49.52\scalebox{0.75}{±1.01} & 58.62\scalebox{0.75}{±6.52} \\
			DataDec~\cite{datadec2023} & \cellcolor{secondplace}{54.23\scalebox{0.75}{±4.16}}  & 63.10\scalebox{0.75}{±2.62} & 58.20\scalebox{0.75}{±4.05} & \cellcolor{secondplace}{69.08\scalebox{0.75}{±4.49}} \\
			ImGKB~\cite{imgkb2023} & 52.83\scalebox{0.75}{±5.45}& \cellcolor{secondplace}{64.93\scalebox{0.75}{±4.44}} & 50.15\scalebox{0.75}{±0.39} & 50.29\scalebox{0.75}{±0.19} \\
			SamGoG & \cellcolor{firstplace}{\bf\color{red}59.49\scalebox{0.75}{±2.68}} & \cellcolor{firstplace}{\bf\color{red}75.13\scalebox{0.75}{±0.90}} & \cellcolor{firstplace}{\bf\color{red}71.70\scalebox{0.75}{±0.79}} & \cellcolor{firstplace}{\bf\color{red}80.05\scalebox{0.75}{±0.23}} \\
			\midrule    
			\multicolumn{5}{c}{Low Imbalance ($\rho_{\text{class}}=7:3$)}\\
			\midrule
			GCN (bb.)~\cite{kipf2017semisupervised} & 43.84\scalebox{0.75}{±7.03} & 58.32\scalebox{0.75}{±1.51} & 49.06\scalebox{0.75}{±2.17} & 61.85\scalebox{0.75}{±3.89} \\
			G\textsuperscript{2}GNN~\cite{g2gnn2022} & 47.86\scalebox{0.75}{±9.03} & 65.65\scalebox{0.75}{±5.41} & 53.11\scalebox{0.75}{±3.97} & 66.02\scalebox{0.75}{±2.08} \\
			TopoImb~\cite{topoimb2022} & 45.90\scalebox{0.75}{±6.41} & 44.63\scalebox{0.75}{±4.82} & 49.75\scalebox{0.75}{±4.81} & 54.35\scalebox{0.75}{±3.13} \\
			DataDec~\cite{datadec2023} & \cellcolor{secondplace}{55.86\scalebox{0.75}{±2.49}} & 64.66\scalebox{0.75}{±2.04} & \cellcolor{secondplace}{57.06\scalebox{0.75}{±5.97}} & \cellcolor{secondplace}{66.45\scalebox{0.75}{±5.38}} \\
			ImGKB~\cite{imgkb2023} & 49.49\scalebox{0.75}{±5.12} & \cellcolor{secondplace}{67.44\scalebox{0.75}{±3.50}} & 47.65\scalebox{0.75}{±0.23} & 48.58\scalebox{0.75}{±2.15} \\
			SamGoG & \cellcolor{firstplace}{\bf\color{red}56.98\scalebox{0.75}{±3.65}} & \cellcolor{firstplace}{\bf\color{red}74.81\scalebox{0.75}{±0.99}} & \cellcolor{firstplace}{\bf\color{red}68.35\scalebox{0.75}{±1.59}} & \cellcolor{firstplace}{\bf\color{red}80.03\scalebox{0.75}{±1.03}} \\
			\midrule
			\multicolumn{5}{c}{High Imbalance ($\rho_{\text{class}}=9:1$)}\\
			\midrule
			GCN (bb.)~\cite{kipf2017semisupervised} & 40.58\scalebox{0.75}{±7.61}  & 42.48\scalebox{0.75}{±2.61} & 45.09\scalebox{0.75}{±0.18} & 54.39\scalebox{0.75}{±5.16} \\
			G\textsuperscript{2}GNN~\cite{g2gnn2022} & 41.78\scalebox{0.75}{±7.61}  & 44.59\scalebox{0.75}{±4.72} & 56.40\scalebox{0.75}{±1.71} & 65.08\scalebox{0.75}{±3.08} \\
			TopoImb~\cite{topoimb2022} & 40.17\scalebox{0.75}{±2.21}  & 36.50\scalebox{0.75}{±1.44} & 46.18\scalebox{0.75}{±3.06} & 52.21\scalebox{0.75}{±3.56} \\
			DataDec~\cite{datadec2023} & \cellcolor{secondplace}{55.41\scalebox{0.75}{±3.37}}  & \cellcolor{secondplace}{65.07\scalebox{0.75}{±2.40}} & \cellcolor{secondplace}{58.58\scalebox{0.75}{±5.07}} & \cellcolor{secondplace}{65.56\scalebox{0.75}{±8.61}} \\
			ImGKB~\cite{imgkb2023} & 45.98\scalebox{0.75}{±8.71} & 58.21\scalebox{0.75}{±10.60} & 46.06\scalebox{0.75}{±2.74} & 51.23\scalebox{0.75}{±5.09} \\
			SamGoG & \cellcolor{firstplace}{\bf\color{red}59.49\scalebox{0.75}{±1.26}} & \cellcolor{firstplace}{\bf\color{red}66.90\scalebox{0.75}{±1.83}} & \cellcolor{firstplace}{\bf\color{red}61.55\scalebox{0.75}{±1.30}} & \cellcolor{firstplace}{\bf\color{red}76.71\scalebox{0.75}{±1.30}} \\
			\bottomrule
		\end{tabular}%
	}
}
\label{tab:main_graph_cls_gcn_acc_all}
\end{minipage}
\end{table}

\begin{table}[!ht]
\vspace{-10pt}
\centering
\caption{Accuracy (\% ± stddev) on datasets with changing size imbalance levels over 10 runs, encoded by GIN / GCN. {\colorbox{firstplace}{\color{red}\bf Red}} and  {\colorbox{secondplace}{Blue}} highlight the best and second-best results respectively.}
\begin{minipage}[t]{0.49\linewidth}  
\centering
\resizebox{\textwidth}{!}{  
	\setlength{\tabcolsep}{0.8em}
	{\footnotesize
		\begin{tabular}{lcccc}
			\toprule
			\textbf{Algorithm} & \textbf{PTC-MR}   & \textbf{D\&D} & \textbf{IMDB-B} & \textbf{REDDIT-B} \\
			\midrule
			Low level $\rho_{\text{size}}$ & 1.60 & 1.56 & 1.64 & 2.79 \\
			\midrule
			GIN (bb.)~\cite{xu2018powerful} & \cellcolor{secondplace}{52.17\scalebox{0.75}{±4.36}} & 65.01\scalebox{0.75}{±0.69} & \cellcolor{secondplace}{66.38\scalebox{0.75}{±4.27}} & 67.41\scalebox{0.75}{±2.32}\\
			SOLT-GNN~\cite{soltgnn2022} & 47.54\scalebox{0.75}{±4.33} & 63.78\scalebox{0.75}{±1.06} & 64.20\scalebox{0.75}{±5.46} & 49.57\scalebox{0.75}{±6.78}\\
			TopoImb~\cite{topoimb2022} & 49.71\scalebox{0.75}{±1.98}  & \cellcolor{secondplace}{65.16\scalebox{0.75}{±1.76}} & 66.00\scalebox{0.75}{±2.41} & \cellcolor{secondplace}{69.47\scalebox{0.75}{±3.70}}\\
			SamGoG & \cellcolor{firstplace}{\bf\color{red}53.86\scalebox{0.75}{±2.12}} & \cellcolor{firstplace}{\bf\color{red}68.22\scalebox{0.75}{±2.40}} & \cellcolor{firstplace}{\bf\color{red}66.82\scalebox{0.75}{±0.85}} & \cellcolor{firstplace}{\bf\color{red}72.20\scalebox{0.75}{±4.24}} \\
			\midrule    
			Mid level $\rho_{\text{size}}$ & 2.90  & 2.53 & 2.11 & 4.92 \\
			\midrule
			GIN (bb.)~\cite{xu2018powerful} & 51.38\scalebox{0.75}{±6.78} & 61.46\scalebox{0.75}{±2.43} & 65.08\scalebox{0.75}{±5.78} & 68.32\scalebox{0.75}{±1.77} \\
			SOLT-GNN~\cite{soltgnn2022} & \cellcolor{secondplace}{53.04\scalebox{0.75}{±3.91}} & 63.33\scalebox{0.75}{±1.86} & \cellcolor{secondplace}{69.38\scalebox{0.75}{±1.23}} & \cellcolor{secondplace}{73.51\scalebox{0.75}{±1.14}}  \\
			TopoImb~\cite{topoimb2022} & 51.59\scalebox{0.75}{±4.30} & \cellcolor{secondplace}{65.99\scalebox{0.75}{±1.25}} & 68.10\scalebox{0.75}{±0.87} & 71.54\scalebox{0.75}{±0.75}  \\
			SamGoG & \cellcolor{firstplace}{\bf\color{red}57.17\scalebox{0.75}{±1.06}} & \cellcolor{firstplace}{\bf\color{red}71.91\scalebox{0.75}{±1.22}} & \cellcolor{firstplace}{\bf\color{red}69.53\scalebox{0.75}{±1.01}} & \cellcolor{firstplace}{\bf\color{red}79.12\scalebox{0.75}{±0.90}} \\
			\midrule
			High level $\rho_{\text{size}}$ & 4.93 & 8.91 & 4.45 & 10.97 \\
			\midrule
			GIN (bb.)~\cite{xu2018powerful} & 48.41\scalebox{0.75}{±7.07} & 60.68\scalebox{0.75}{±6.89} & 62.60\scalebox{0.75}{±3.82} & 67.41\scalebox{0.75}{±2.23} \\
			SOLT-GNN~\cite{soltgnn2022} & 51.74\scalebox{0.75}{±5.25} & \cellcolor{secondplace}{64.97\scalebox{0.75}{±3.24}} & 65.03\scalebox{0.75}{±4.12} & 60.24\scalebox{0.75}{±2.11} \\
			TopoImb~\cite{topoimb2022} & \cellcolor{secondplace}{51.96\scalebox{0.75}{±1.16}} & 64.16\scalebox{0.75}{±2.96} & \cellcolor{secondplace}{66.75\scalebox{0.75}{±0.91}} & \cellcolor{secondplace}{69.14\scalebox{0.75}{±4.83}} \\
			SamGoG & \cellcolor{firstplace}{\bf\color{red}59.13\scalebox{0.75}{±1.32}} & \cellcolor{firstplace}{\bf\color{red}69.15\scalebox{0.75}{±1.87}} & \cellcolor{firstplace}{\bf\color{red}67.13\scalebox{0.75}{±1.59}} & \cellcolor{firstplace}{\bf\color{red}72.31\scalebox{0.75}{±2.07}} \\
			\bottomrule
		\end{tabular}%
	}
}
\label{tab:main_graph_topo_gin_acc_all}
\end{minipage}
\hfill 
\begin{minipage}[t]{0.49\linewidth}  
\centering
\resizebox{\textwidth}{!}{  
	\setlength{\tabcolsep}{0.8em}
	{\footnotesize
		\begin{tabular}{lcccc}
			\toprule
			\textbf{Algorithm} & \textbf{PTC-MR}   & \textbf{D\&D} & \textbf{IMDB-B} & \textbf{REDDIT-B} \\
			\midrule
			Low level $\rho_{\text{size}}$ & 1.60 & 1.56 & 1.64 & 2.79 \\
			\midrule
			GCN (bb.)~\cite{kipf2017semisupervised} & \cellcolor{secondplace}{51.59\scalebox{0.75}{±7.07}} & 64.82\scalebox{0.75}{±0.75} & \cellcolor{secondplace}{69.15\scalebox{0.75}{±1.44}} & 53.44\scalebox{0.75}{±1.78} \\
			SOLT-GNN~\cite{soltgnn2022} & 51.45\scalebox{0.75}{±3.13} & \cellcolor{secondplace}{65.69\scalebox{0.75}{±2.36}} & 68.23\scalebox{0.75}{±2.48} & 37.58\scalebox{0.75}{±3.04} \\
			TopoImb~\cite{topoimb2022} & 46.74\scalebox{0.75}{±1.57} & 56.18\scalebox{0.75}{±2.10} & 68.20\scalebox{0.75}{±0.70} & \cellcolor{secondplace}{56.94\scalebox{0.75}{±3.23}}  \\
			SamGoG & \cellcolor{firstplace}{\bf\color{red}52.83\scalebox{0.75}{±0.90}} & \cellcolor{firstplace}{\bf\color{red}71.06\scalebox{0.75}{±3.54}} & \cellcolor{firstplace}{\bf\color{red}70.53\scalebox{0.75}{±0.87}} & \cellcolor{firstplace}{\bf\color{red}72.60\scalebox{0.75}{±1.72}} \\
			\midrule    
			Mid level $\rho_{\text{size}}$ & 2.90  & 2.53 & 2.11 & 4.92 \\
			\midrule
			GCN (bb.)~\cite{kipf2017semisupervised} & 50.00\scalebox{0.75}{±5.87}  & 63.99\scalebox{0.75}{±2.16} & 67.95\scalebox{0.75}{±2.82} & 67.96\scalebox{0.75}{±0.89} \\
			SOLT-GNN~\cite{soltgnn2022} & \cellcolor{secondplace}{56.23\scalebox{0.75}{±0.84}}  & 62.78\scalebox{0.75}{±1.33} & \cellcolor{secondplace}{70.25\scalebox{0.75}{±1.15}} & 61.95\scalebox{0.75}{±5.55}  \\
			TopoImb~\cite{topoimb2022} & 54.13\scalebox{0.75}{±4.62} & \cellcolor{secondplace}{64.73\scalebox{0.75}{±7.09}} & 68.75\scalebox{0.75}{±0.76} & \cellcolor{secondplace}{69.12\scalebox{0.75}{±0.52}} \\
			SamGoG & \cellcolor{firstplace}{\bf\color{red}57.03\scalebox{0.75}{±1.46}} & \cellcolor{firstplace}{\bf\color{red}72.80\scalebox{0.75}{±2.53}} & \cellcolor{firstplace}{\bf\color{red}70.54\scalebox{0.75}{±0.40}} & \cellcolor{firstplace}{\bf\color{red}79.08\scalebox{0.75}{±1.36}} \\
			\midrule
			High level $\rho_{\text{size}}$ & 4.93 & 8.91 & 4.45 & 10.97 \\
			\midrule
			GCN (bb.)~\cite{kipf2017semisupervised} & 49.93\scalebox{0.75}{±5.90}  & 60.98\scalebox{0.75}{±6.71} & 64.88\scalebox{0.75}{±2.02} & 66.38\scalebox{0.75}{±0.46} \\
			SOLT-GNN~\cite{soltgnn2022} & \cellcolor{secondplace}{54.78\scalebox{0.75}{±6.03}} & 63.21\scalebox{0.75}{±3.40} & \cellcolor{firstplace}{\bf\color{red}69.80\scalebox{0.75}{±2.07}} & 67.05\scalebox{0.75}{±2.54} \\
			TopoImb~\cite{topoimb2022} & 51.81\scalebox{0.75}{±1.26} & \cellcolor{secondplace}{69.66\scalebox{0.75}{±1.92}} & 66.60\scalebox{0.75}{±0.91} & \cellcolor{secondplace}{69.09\scalebox{0.75}{±1.00}} \\
			SamGoG & \cellcolor{firstplace}{\bf\color{red}59.71\scalebox{0.75}{±1.46}} & \cellcolor{firstplace}{\bf\color{red}74.65\scalebox{0.75}{±1.45}} & \cellcolor{secondplace}{69.38\scalebox{0.75}{±0.30}} & \cellcolor{firstplace}{\bf\color{red}77.29\scalebox{0.75}{±1.26}} \\
			\bottomrule
		\end{tabular}%
	}
}
\vspace{-30pt}
\label{tab:main_graph_topo_gcn_acc_all}
\end{minipage}
\end{table}

\vspace{-0.7em}
\subsection{Performance Analysis}
\vspace{-0.7em}
To answer \textbf{RQ1}, as shown in~\cref{tab:main_graph_cls_gcn_acc_all,tab:main_graph_topo_gcn_acc_all}, SamGoG achieves state-of-the-art performance on 23/24 for both imbalance types, surpassing the second-best baseline by {\bf 5.07\%} (class) and {\bf 4.13\%} (size) in average accuracy. 
To answer \textbf{RQ2}, balanced accuracy and macro-F1 scores in Appendix~\ref{sec:addiexp} confirm consistent improvements across minority classes. For graph size imbalance, as shown in~\cref{fig:exp2}, SamGoG simultaneously enhances accuracy on both head- and tail-size graphs compared to vanilla GIN~\cite{xu2018powerful}, demonstrating effective mitigation of graph size imbalance. To confirm performance does not rely on the specific downstream model, we adopt GraphSHA~\cite{graphsha} as an additional model on D\&D dataset. Results in Appendix~\ref{sec:graphsha} demonstrates SamGoG's potential to leverage various downstream models.
\cref{fig:samplet} reports the relationship between sample size $t$ and accuracy.
\begin{figure*}[t]
	\vspace{-10pt}
	\begin{minipage}[c]{0.5\textwidth}
		\centering
		\includegraphics[width=\linewidth]{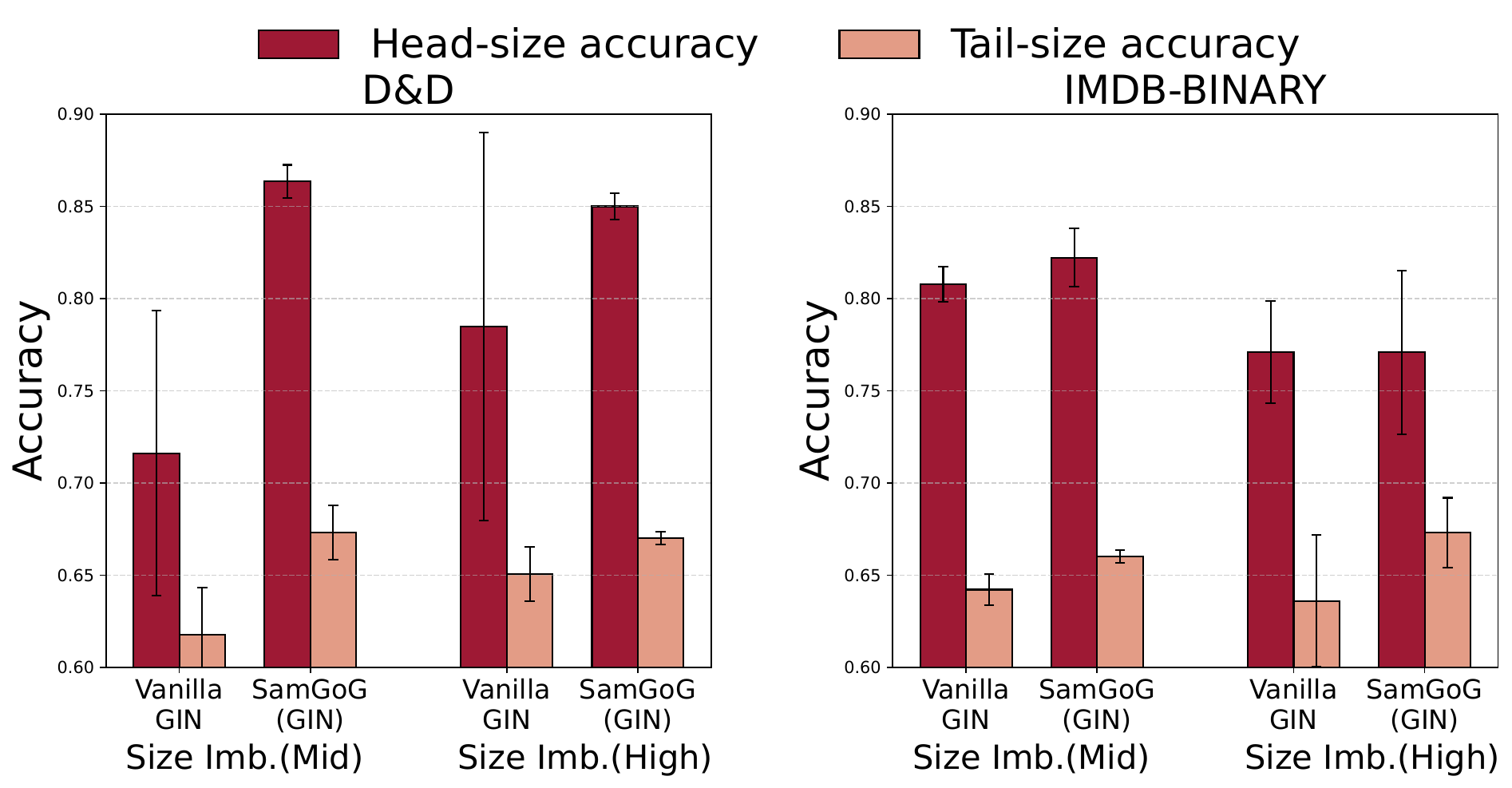}
		\caption{Test accuracy (± stddev) on head-/tail-size graphs by vanilla GIN and SamGoG with GIN as encoder.}
		\label{fig:exp2}
	\end{minipage}
	\begin{minipage}[c]{0.5\textwidth}
		\centering
		\includegraphics[width=\linewidth]{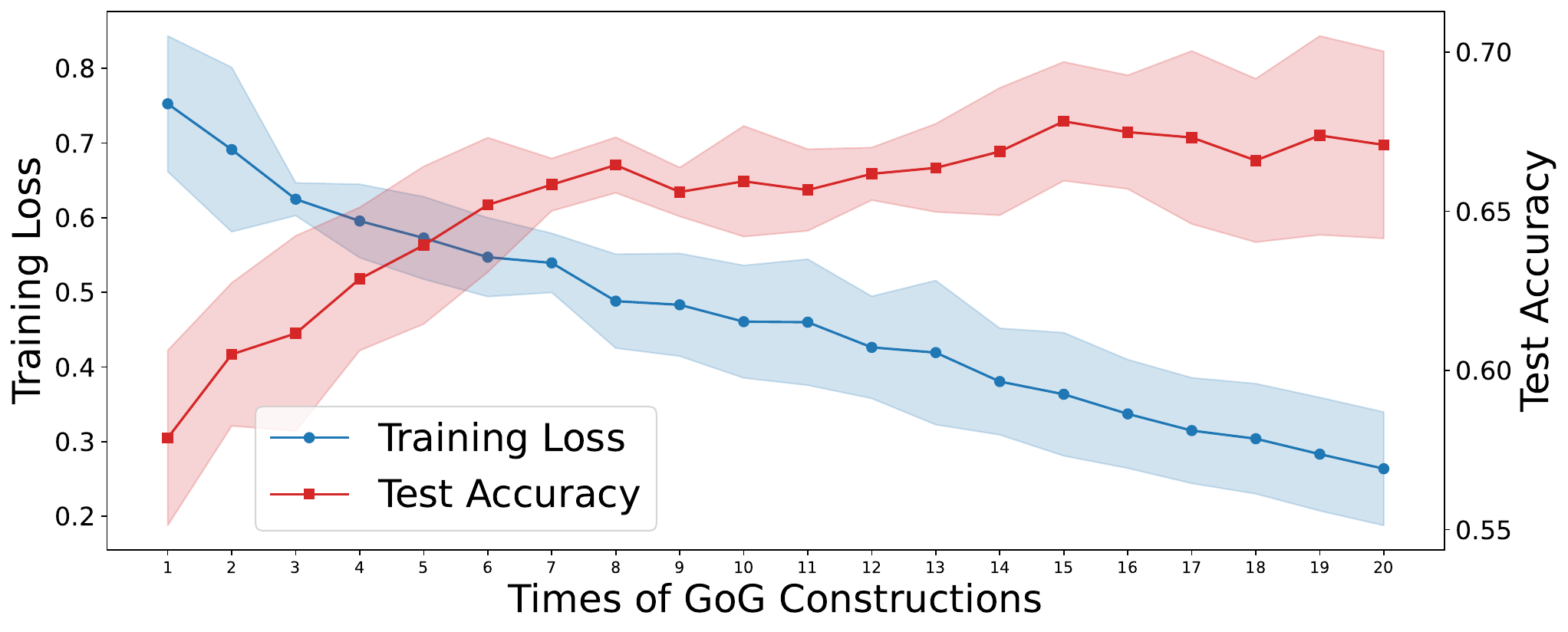}
		\caption{Train loss and test accuracy (± stddev) of SamGoG(GCN) on 20 times of GoG Constructions and downstream model training epochs on mid-level class imbalance split of D\&D dataset.}
		\label{fig:samplet}
	\end{minipage}
	\vspace{-12pt}
\end{figure*}%
\begin{figure*}[t]
	\vspace{-10pt}
	\begin{minipage}[c]{0.25\textwidth}
		\centering
		\includegraphics[width=\linewidth]{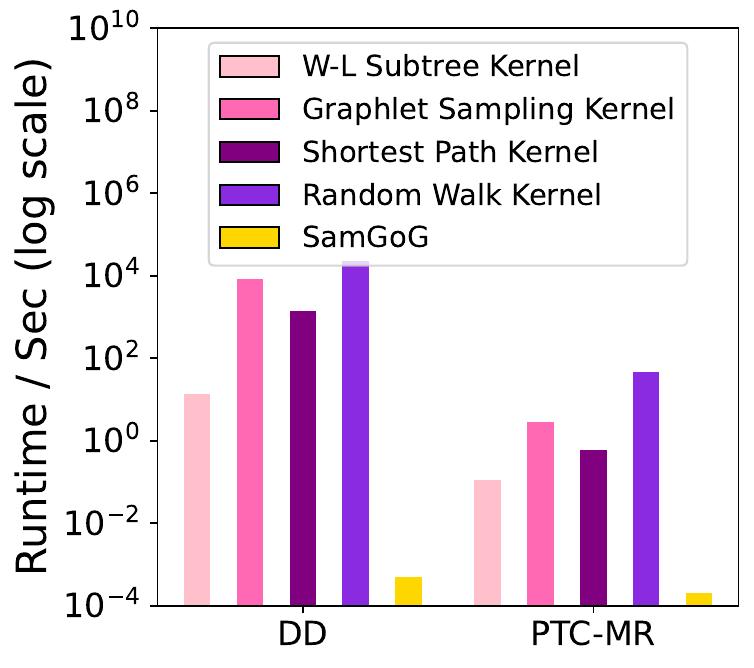}
		\caption{Runtime (seconds) of four kernel methods and SamGoG on D\&D and PTC-MR datasets.}
		\label{fig:simitime}
	\end{minipage}
	\begin{minipage}[c]{0.75\textwidth}
		\centering
		\includegraphics[width=\linewidth]{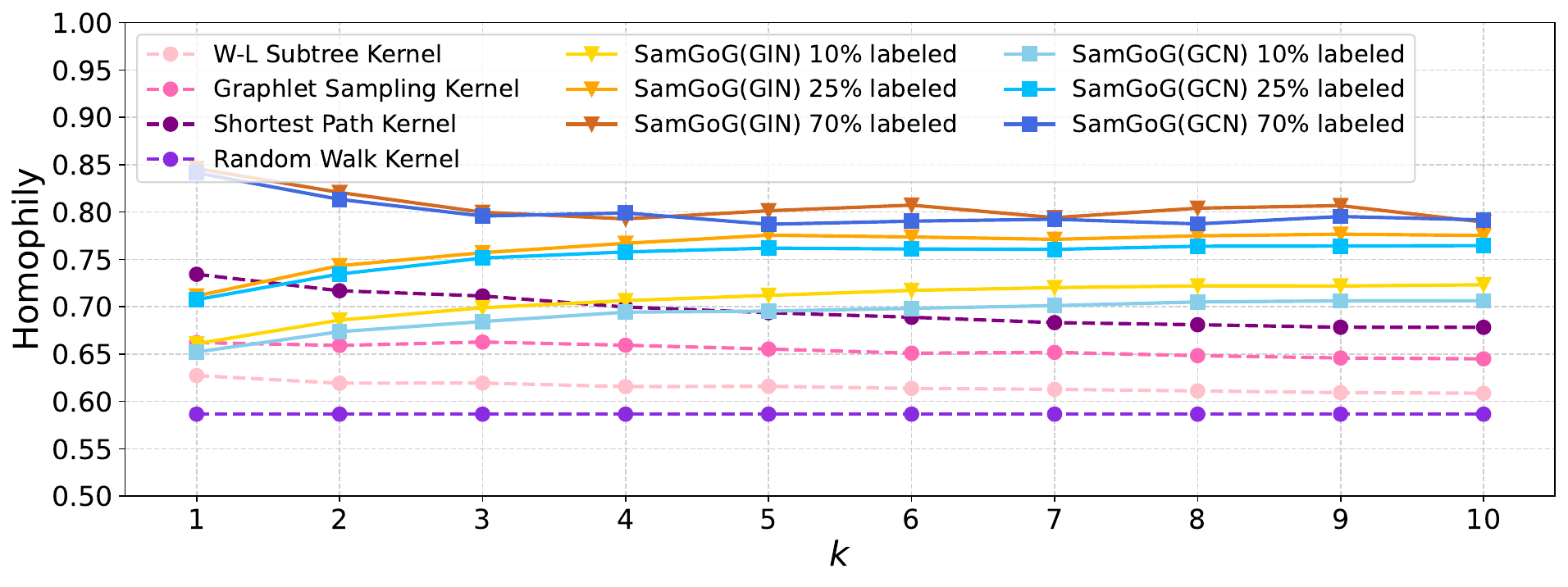}
		\caption{Edge homophily with different average degree $k$: SamGoG with GIN or GCN as encoder on 10\%/25\%/70\% labeled data preserve competitive edge and node homophily.}
		\label{fig:exp3}
	\end{minipage}
	\vspace{-10pt}
\end{figure*}%
\begin{figure*}[t]
	\begin{minipage}[c]{0.3\textwidth}
		\centering
		\includegraphics[width=\linewidth]{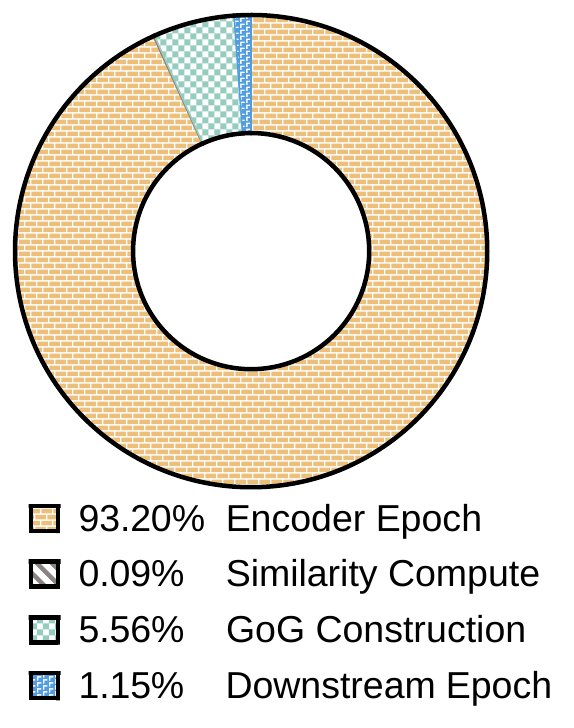}
		\vspace{-2.0em}
		\caption{Runtime Proportion of SamGoG running on ogbg-molhiv dataset.}
		\label{fig:molhivtime}
	\end{minipage}
	\begin{minipage}[c]{0.7\textwidth}
		\centering
		\captionof{table}{Accuracy (\% ± stddev), Macro-F1 (\% ± stddev), average preprocessing time (second), and average epoch time (second) of graph classification on ogbg-molhiv dataset with low-level class-imbalanced split over 10 runs. ``OOM'' denotes out of memory or time limit.}
		\label{tab:ogbg}
		\footnotesize
		\setlength{\tabcolsep}{4pt}
		\begin{tabular}{lcccc}
			\toprule
			\textbf{Algorithm} & \textbf{Acc.} & \textbf{Macro-F1} & \textbf{\makecell[c]{Preprocessing\\Time (s)}}& \textbf{\makecell[c]{Epoch\\Time (s)}}\\
			\midrule
			\multicolumn{5}{c}{Low Class Imbalance ($\rho_{\text{class}}=20$)}\\
			\midrule
			G\textsuperscript{2}GNN~\cite{g2gnn2022} & OOM & OOM & OOM & OOM \\
			TopoImb~\cite{topoimb2022} & OOM  & OOM  & OOM  & OOM \\
			DataDec~\cite{datadec2023} & OOM  & OOM  & OOM  & OOM \\
			SOLT-GNN~\cite{soltgnn2022}  & OOM  & OOM  & OOM  & OOM \\
			ImGKB~\cite{imgkb2023} &
			\cellcolor{secondplace}{67.50\scalebox{0.75}{±2.70}} &
			\cellcolor{secondplace}{65.87\scalebox{0.75}{±3.81}} &
			\cellcolor{secondplace}{{12.83}} & \cellcolor{secondplace}{{3.77}} \\
			\makecell[l]{SamGoG} & \cellcolor{firstplace}{\bf\color{red}71.67\scalebox{0.75}{±0.93}} &
			\cellcolor{firstplace}{\bf\color{red}71.05\scalebox{0.75}{±1.12}} &
		\cellcolor{firstplace}{\bf\color{red}{0.03}} & 		\cellcolor{firstplace}{\bf\color{red}{3.58}} \\
			\bottomrule
		\end{tabular}
	\end{minipage}
		\begin{minipage}{0.33\textwidth}
		\centering
		\includegraphics[width=\linewidth]{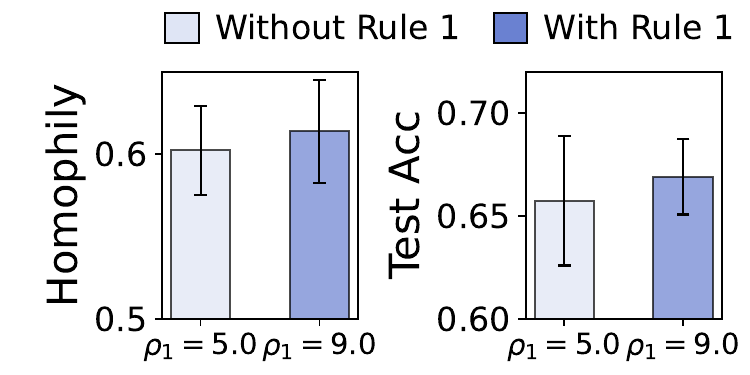}
		\caption{Edge Homophily (± stddev) and test accuracy (± standard deviation) w/o and w/~\cref{rule:rule1}.}
		\label{fig:ablar1}
	\end{minipage}
	\begin{minipage}{0.33\textwidth}
		\centering
		\includegraphics[width=\linewidth]{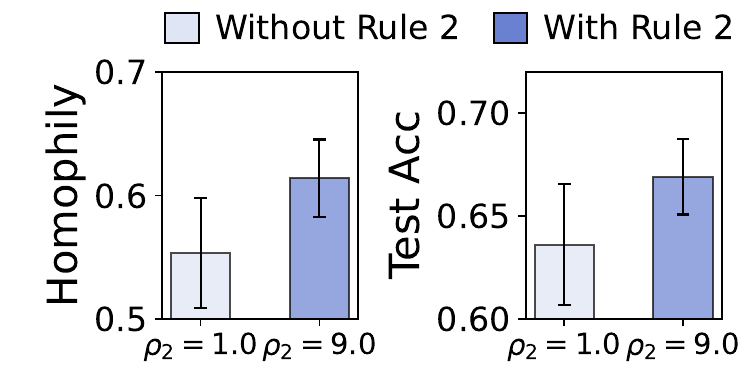}
		\caption{Edge Homophily (± stddev) and test accuracy (± standard deviation) w/o and w/~\cref{rule:rule2}.}
		\label{fig:ablar2}
	\end{minipage}
	\begin{minipage}{0.33\textwidth}
		\centering
		\includegraphics[width=\linewidth]{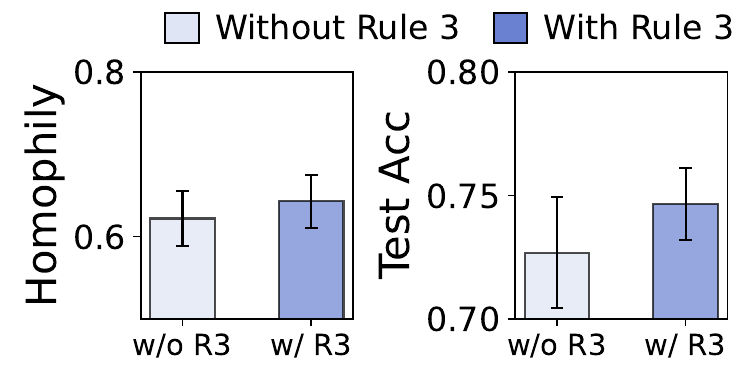}
		\caption{Edge Homophily (± stddev) and test accuracy (± standard deviation) w/o and w/~\cref{rule:rule3}.}
		\label{fig:ablar3}
	\end{minipage}
\end{figure*}%
\vspace{-0.7em}
\subsection{Efficiency Studies}
\vspace{-0.7em}
To answer \textbf{RQ4}, we measure the time needed for two key steps in the GoG pipeline: calculating the pairwise graph similarity matrix and building the GoGs by sampling. We test this on the TUDataset (about 2,000 graphs) and the ogbg-molhiv dataset (over 41,000 graphs).
{As shown in~\cref{fig:simitime}, SamGoG achieves remarkable efficiency in pairwise similarity computation, completing the task in milliseconds on TUDataset series (about 1,000\textasciitilde2,000 graphs). This represents a significant improvement over graph kernel methods, which require seconds to minutes for preprocessing at comparable scales. SOLT-GNN~\cite{soltgnn2022} requires 2,276 seconds for preprocessing and 500 epochs training on REDDIT-B, while SamGoG with an equally-sized model completes in 340 seconds, achieving a 6.7$\times$ speedup. 
The efficiency  advantage is further demonstrated in~\cref{fig:molhivtime,tab:ogbg} on the largest ogbg-molhiv dataset. SamGoG achieves SOTA accuracy and macro-F1 while maintaining the shortest preprocessing time. 
In SamGoG, the additional overhead of GoG construction and downstream model training per epoch accounts for only 7\% of the total epoch time, highlighting its exceptional efficiency . These results collectively underscore SamGoG's superior performance in handling large-scale imbalanced graph datasets.
}
	
\vspace{-0.7em}
\subsection{Ablation Studies}
\vspace{-0.7em}
In this section, we perform systematic ablation studies to investigate the individual contributions of key components in SamGoG. As evidenced in the class imbalance experiment (Table~\ref{tab:main_graph_cls_gcn_acc_all}), SamGoG significantly outperforms G\textsuperscript{2}GNN~\cite{g2gnn2022}, a result we attribute to its superior edge homophily compared to the kernel-based methods discussed in~\cref{fig:exp3}. This also answer \textbf{RQ3}. To quantify the impact of our proposed node degree rules (Rules~\ref{rule:rule1}-\ref{rule:rule3}), we systematically evaluate three model variants by removing each rule independently. 
Specifically, we assess Rules~\ref{rule:rule1} and~\ref{rule:rule2} under the class-imbalanced partition of the D\&D dataset, while Rule~\ref{rule:rule3} is analyzed using the size-imbalanced partition. All experiments employ SamGoG with GCN as the backbone model, with mean and standard deviation metrics computed over 10 independent runs for both edge homophily and test accuracy. The ablation results (Figures~\ref{fig:ablar1}, \ref{fig:ablar2}, and \ref{fig:ablar3}) demonstrate that the integration of all three rules yields a consistent improvement of several percentage points in both edge homophily synchronization and classification accuracy, thereby empirically validating their effectiveness.

	\vspace{-0.7em}
\section{Conclusion}
\vspace{-0.7em}
This work introduces SamGoG, a sampling-based graph-of-graphs framework that effectively tackles class and graph size imbalance in graph classification. SamGoG enhances edge homophily of GoG by learnable pairwise graph similarity and adaptive GoG node degree allocation. The importance-based sampling mechanism ensures efficient construction of multiple GoGs and seamless integration with various downstream GNNs. SamGoG outperforms existing methods in both accuracy and efficiency.
	\clearpage
	\newpage
	\bibliography{refs}

\begin{thebibliography}{10}

\bibitem{10.1145/3308558.3313488}
Wenqi Fan, Yao Ma, Qing Li, Yuan He, Yihong~Eric Zhao, Jiliang Tang, and Dawei
  Yin.
\newblock Graph neural networks for social recommendation.
\newblock In Ling Liu, Ryen~W. White, Amin Mantrach, Fabrizio Silvestri,
  Julian~J. McAuley, Ricardo Baeza{-}Yates, and Leila Zia, editors, {\em The
  World Wide Web Conference, {WWW} 2019}, pages 417--426. {ACM}.

\bibitem{jha2022prediction}
Kanchan Jha, Sriparna Saha, and Hiteshi Singh.
\newblock Prediction of protein--protein interaction using graph neural
  networks.
\newblock {\em Scientific Reports}, 12(1):8360, 2022.

\bibitem{Prot2Text}
Hadi Abdine, Michail Chatzianastasis, Costas Bouyioukos, and Michalis
  Vazirgiannis.
\newblock Prot2text: Multimodal protein's function generation with gnns and
  transformers.
\newblock pages 10757--10765.

\bibitem{kipf2017semisupervised}
Thomas~N. Kipf and Max Welling.
\newblock Semi-supervised classification with graph convolutional networks.
\newblock In {\em 5th International Conference on Learning Representations,
  {ICLR} 2017, Conference Track Proceedings}. OpenReview.net.

\bibitem{GAT2018}
Petar Velickovic, Guillem Cucurull, Arantxa Casanova, Adriana Romero, Pietro
  Li{\`{o}}, and Yoshua Bengio.
\newblock Graph attention networks.
\newblock In {\em 6th International Conference on Learning Representations,
  {ICLR} 2018, Conference Track Proceedings}. OpenReview.net.

\bibitem{liu2024classimbalancedgraphlearningclass}
Zhining Liu, Ruizhong Qiu, Zhichen Zeng, Hyunsik Yoo, David Zhou, Zhe Xu, Yada
  Zhu, Kommy Weldemariam, Jingrui He, and Hanghang Tong.
\newblock Class-imbalanced graph learning without class rebalancing.

\bibitem{pan2013graph}
Shirui Pan and Xingquan Zhu.
\newblock Graph classification with imbalanced class distributions and noise.
\newblock In Francesca Rossi, editor, {\em {IJCAI} 2013, Proceedings of the
  23rd International Joint Conference on Artificial Intelligence}, pages
  1586--1592. {IJCAI/AAAI}.

\bibitem{topoimb2022}
Tianxiang Zhao, Dongsheng Luo, Xiang Zhang, and Suhang Wang.
\newblock Topoimb: Toward topology-level imbalance in learning from graphs.
\newblock In Bastian Rieck and Razvan Pascanu, editors, {\em Learning on Graphs
  Conference, LoG 2022}, volume 198 of {\em Proceedings of Machine Learning
  Research}, page~37. {PMLR}.

\bibitem{soltgnn2022}
Zemin Liu, Qiheng Mao, Chenghao Liu, Yuan Fang, and Jianling Sun.
\newblock On size-oriented long-tailed graph classification of graph neural
  networks.
\newblock In Fr{\'{e}}d{\'{e}}rique Laforest, Rapha{\"{e}}l Troncy, Elena
  Simperl, Deepak Agarwal, Aristides Gionis, Ivan Herman, and Lionel
  M{\'{e}}dini, editors, {\em {WWW} '22: The {ACM} Web Conference 2022}, pages
  1506--1516. {ACM}.

\bibitem{chawla2002smote}
Nitesh~V. Chawla, Kevin~W. Bowyer, Lawrence~O. Hall, and W.~Philip Kegelmeyer.
\newblock {SMOTE:} synthetic minority over-sampling technique.
\newblock {\em J. Artif. Intell. Res.}, 16:321--357, 2002.

\bibitem{drummond2003c4}
Chris Drummond, Robert~C Holte, et~al.
\newblock C4. 5, class imbalance, and cost sensitivity: why under-sampling
  beats over-sampling.
\newblock In {\em Workshop on learning from imbalanced datasets II}, volume~11,
  2003.

\bibitem{wu2021graphmixupimprovingclassimbalancednode}
Lirong Wu, Haitao Lin, Zhangyang Gao, Cheng Tan, and Stan~Z. Li.
\newblock Graphmixup: Improving class-imbalanced node classification on graphs
  by self-supervised context prediction, 2021.

\bibitem{imbgnn2024}
Wei Xu, Pengkun Wang, Zhe Zhao, Binwu Wang, Xu~Wang, and Yang Wang.
\newblock When imbalance meets imbalance: Structure-driven learning for
  imbalanced graph classification.
\newblock In Tat{-}Seng Chua, Chong{-}Wah Ngo, Ravi Kumar, Hady~W. Lauw, and
  Roy~Ka{-}Wei Lee, editors, {\em Proceedings of the {ACM} on Web Conference,
  {WWW} 2024}, pages 905--913. {ACM}.

\bibitem{imgkb2023}
Hui Tang and Xun Liang.
\newblock Where to find fascinating inter-graph supervision: Imbalanced graph
  classification with kernel information bottleneck.
\newblock In Abdulmotaleb El{-}Saddik, Tao Mei, Rita Cucchiara, Marco Bertini,
  Diana Patricia~Tobon Vallejo, Pradeep~K. Atrey, and M.~Shamim Hossain,
  editors, {\em Proceedings of the 31st {ACM} International Conference on
  Multimedia, {MM} 2023}, pages 3240--3249. {ACM}.

\bibitem{g2gnn2022}
Yu~Wang, Yuying Zhao, Neil Shah, and Tyler Derr.
\newblock Imbalanced graph classification via graph-of-graph neural networks.
\newblock In Mohammad~Al Hasan and Li~Xiong, editors, {\em Proceedings of the
  31st {ACM} International Conference on Information {\&} Knowledge Management,
  2022}, pages 2067--2076. {ACM}.

\bibitem{wlstkernel}
Nino Shervashidze, Pascal Schweitzer, Erik~Jan van Leeuwen, Kurt Mehlhorn, and
  Karsten~M. Borgwardt.
\newblock Weisfeiler-lehman graph kernels.
\newblock {\em J. Mach. Learn. Res.}, 12:2539--2561, 2011.

\bibitem{xu2018powerful}
Keyulu Xu, Weihua Hu, Jure Leskovec, and Stefanie Jegelka.
\newblock How powerful are graph neural networks?
\newblock In {\em 7th International Conference on Learning Representations,
  {ICLR} 2019, Conference Track Proceedings}. OpenReview.net.

\bibitem{spkernel}
Karsten~M. Borgwardt and Hans{-}Peter Kriegel.
\newblock Shortest-path kernels on graphs.
\newblock In {\em Proceedings of the 5th {IEEE} International Conference on
  Data Mining {(ICDM} 2005)}, pages 74--81. {IEEE} Computer Society.

\bibitem{ptcmr}
Yunsheng Bai, Hao Ding, Yang Qiao, Agustin Marinovic, Ken Gu, Ting Chen, Yizhou
  Sun, and Wei Wang.
\newblock Unsupervised inductive graph-level representation learning via
  graph-graph proximity.

\bibitem{sutherland2003spline}
Jeffrey~J. Sutherland, Lee~A. O'Brien, and Donald~F. Weaver.
\newblock Spline-fitting with a genetic algorithm: {A} method for developing
  classification structure-activity relationships.
\newblock {\em J. Chem. Inf. Comput. Sci.}, 43(6):1906--1915, 2003.

\bibitem{iglbench}
Jiawen Qin, Haonan Yuan, Qingyun Sun, Lyujin Xu, Jiaqi Yuan, Pengfeng Huang,
  Zhaonan Wang, Xingcheng Fu, Hao Peng, Jianxin Li, and Philip~S. Yu.
\newblock Igl-bench: Establishing the comprehensive benchmark for imbalanced
  graph learning.

\bibitem{10.5555/2969442.2969488}
David Duvenaud, Dougal Maclaurin, Jorge Aguilera{-}Iparraguirre, Rafael
  G{\'{o}}mez{-}Bombarelli, Timothy Hirzel, Al{\'{a}}n Aspuru{-}Guzik, and
  Ryan~P. Adams.
\newblock Convolutional networks on graphs for learning molecular fingerprints.
\newblock In Corinna Cortes, Neil~D. Lawrence, Daniel~D. Lee, Masashi Sugiyama,
  and Roman Garnett, editors, {\em Advances in Neural Information Processing
  Systems 28: Annual Conference on Neural Information Processing Systems 2015},
  pages 2224--2232.

\bibitem{10.1145/2623330.2623643}
Jingchao Ni, Hanghang Tong, Wei Fan, and Xiang Zhang.
\newblock Inside the atoms: ranking on a network of networks.
\newblock In Sofus~A. Macskassy, Claudia Perlich, Jure Leskovec, Wei Wang, and
  Rayid Ghani, editors, {\em The 20th {ACM} {SIGKDD} International Conference
  on Knowledge Discovery and Data Mining, {KDD} '14, 2014}, pages 1356--1365.
  {ACM}.

\bibitem{gu2021modelingmultiscaledatanetwork}
Shawn Gu, Meng Jiang, Pietro~Hiram Guzzi, and Tijana Milenkovic.
\newblock Modeling multi-scale data via a network of networks, 2022.

\bibitem{Wang_2020}
Hanchen Wang, Defu Lian, Ying Zhang, Lu~Qin, and Xuemin Lin.
\newblock Gognn: Graph of graphs neural network for predicting structured
  entity interactions.
\newblock In Christian Bessiere, editor, {\em Proceedings of the Twenty-Ninth
  International Joint Conference on Artificial Intelligence, {IJCAI} 2020},
  pages 1317--1323. ijcai.org.

\bibitem{simba}
Jiawen Qin, Pengfeng Huang, Qingyun Sun, Cheng Ji, Xingcheng Fu, and Jianxin
  Li.
\newblock Graph size-imbalanced learning with energy-guided structural
  smoothing.
\newblock In Wolfgang Nejdl, S{\"{o}}ren Auer, Meeyoung Cha, Marie{-}Francine
  Moens, and Marc Najork, editors, {\em Proceedings of the Eighteenth {ACM}
  International Conference on Web Search and Data Mining, {WSDM} 2025}, pages
  457--465. {ACM}.

\bibitem{zhu2020beyond}
Jiong Zhu, Yujun Yan, Lingxiao Zhao, Mark Heimann, Leman Akoglu, and Danai
  Koutra.
\newblock Beyond homophily in graph neural networks: Current limitations and
  effective designs.

\bibitem{pei2020geomgcngeometricgraphconvolutional}
Hongbin Pei, Bingzhe Wei, Kevin~Chen{-}Chuan Chang, Yu~Lei, and Bo~Yang.
\newblock Geom-gcn: Geometric graph convolutional networks.

\bibitem{lim2021newbenchmarkslearningnonhomophilous}
Derek Lim, Xiuyu Li, Felix Hohne, and Ser{-}Nam Lim.
\newblock New benchmarks for learning on non-homophilous graphs, 2021.

\bibitem{10.1145/3340531.3411910}
Zemin Liu, Wentao Zhang, Yuan Fang, Xinming Zhang, and Steven C.~H. Hoi.
\newblock Towards locality-aware meta-learning of tail node embeddings on
  networks.
\newblock In Mathieu d'Aquin, Stefan Dietze, Claudia Hauff, Edward Curry, and
  Philippe Cudr{\'{e}}{-}Mauroux, editors, {\em {CIKM} '20: The 29th {ACM}
  International Conference on Information and Knowledge Management, 2020},
  pages 975--984. {ACM}.

\bibitem{tailgnn}
Zemin Liu, Trung{-}Kien Nguyen, and Yuan Fang.
\newblock Tail-gnn: Tail-node graph neural networks.
\newblock In Feida Zhu, Beng~Chin Ooi, and Chunyan Miao, editors, {\em {KDD}
  '21: The 27th {ACM} {SIGKDD} Conference on Knowledge Discovery and Data
  Mining}, pages 1109--1119. {ACM}, 2021.

\bibitem{elkan2001foundations}
Charles Elkan.
\newblock The foundations of cost-sensitive learning.
\newblock In Bernhard Nebel, editor, {\em Proceedings of the Seventeenth
  International Joint Conference on Artificial Intelligence, {IJCAI} 2001},
  pages 973--978. Morgan Kaufmann.

\bibitem{ling2008cost}
Abouzar Choubineh, David~A. Wood, and Zahak Choubineh.
\newblock Applying separately cost-sensitive learning and fisher's discriminant
  analysis to address the class imbalance problem: {A} case study involving a
  virtual gas pipeline {SCADA} system.
\newblock {\em Int. J. Crit. Infrastructure Prot.}, 29:100357, 2020.

\bibitem{zhou2006training}
Zhi{-}Hua Zhou and Xu{-}Ying Liu.
\newblock Training cost-sensitive neural networks with methods addressing the
  class imbalance problem.
\newblock {\em {IEEE} Trans. Knowl. Data Eng.}, 18(1):63--77, 2006.

\bibitem{datadec2023}
Chunhui Zhang, Chao Huang, Yijun Tian, Qianlong Wen, Zhongyu Ouyang, Youhuan
  Li, Yanfang Ye, and Chuxu Zhang.
\newblock When sparsity meets contrastive models: Less graph data can bring
  better class-balanced representations.
\newblock In Andreas Krause, Emma Brunskill, Kyunghyun Cho, Barbara Engelhardt,
  Sivan Sabato, and Jonathan Scarlett, editors, {\em International Conference
  on Machine Learning, {ICML} 2023}, volume 202 of {\em Proceedings of Machine
  Learning Research}, pages 41133--41150. {PMLR}.

\bibitem{tudatasets}
Christopher Morris, Nils~M. Kriege, Franka Bause, Kristian Kersting, Petra
  Mutzel, and Marion Neumann.
\newblock Tudataset: {A} collection of benchmark datasets for learning with
  graphs, 2020.

\bibitem{ogbbenchmark}
Weihua Hu, Matthias Fey, Marinka Zitnik, Yuxiao Dong, Hongyu Ren, Bowen Liu,
  Michele Catasta, and Jure Leskovec.
\newblock Open graph benchmark: Datasets for machine learning on graphs.

\bibitem{randomwalkkernel}
Tam{\'{a}}s Horv{\'{a}}th, Thomas G{\"{a}}rtner, and Stefan Wrobel.
\newblock Cyclic pattern kernels for predictive graph mining.
\newblock In Won Kim, Ron Kohavi, Johannes Gehrke, and William DuMouchel,
  editors, {\em Proceedings of the Tenth {ACM} {SIGKDD} International
  Conference on Knowledge Discovery and Data Mining, 2004}, pages 158--167.
  {ACM}, 2004.

\bibitem{gsamk}
Natasa Przulj.
\newblock Biological network comparison using graphlet degree distribution.
\newblock {\em Bioinform.}, 23(2):177--183, 2007.

\bibitem{graphsha}
Wen{-}Zhi Li, Chang{-}Dong Wang, Hui Xiong, and Jian{-}Huang Lai.
\newblock Graphsha: Synthesizing harder samples for class-imbalanced node
  classification.
\newblock In Ambuj~K. Singh, Yizhou Sun, Leman Akoglu, Dimitrios Gunopulos,
  Xifeng Yan, Ravi Kumar, Fatma Ozcan, and Jieping Ye, editors, {\em
  Proceedings of the 29th {ACM} {SIGKDD} Conference on Knowledge Discovery and
  Data Mining, {KDD} 2023}, pages 1328--1340. {ACM}.

\bibitem{imdbb}
Chen Cai and Yusu Wang.
\newblock A simple yet effective baseline for non-attribute graph
  classification, 2018.

\bibitem{redditb}
Pinar Yanardag and S.~V.~N. Vishwanathan.
\newblock Deep graph kernels.
\newblock In Longbing Cao, Chengqi Zhang, Thorsten Joachims, Geoffrey~I. Webb,
  Dragos~D. Margineantu, and Graham Williams, editors, {\em Proceedings of the
  21th {ACM} {SIGKDD} International Conference on Knowledge Discovery and Data
  Mining, 2015}, pages 1365--1374. {ACM}.

\bibitem{wu2018moleculenet}
Zhenqin Wu, Bharath Ramsundar, Evan~N. Feinberg, Joseph Gomes, Caleb Geniesse,
  Aneesh~S. Pappu, Karl Leswing, and Vijay~S. Pande.
\newblock Moleculenet: {A} benchmark for molecular machine learning.
\newblock {\em CoRR}, abs/1703.00564, 2017.

\bibitem{pytorch}
Adam Paszke, Sam Gross, Francisco Massa, Adam Lerer, James Bradbury, Gregory
  Chanan, Trevor Killeen, Zeming Lin, Natalia Gimelshein, Luca Antiga, Alban
  Desmaison, Andreas K{\"{o}}pf, Edward~Z. Yang, Zachary DeVito, Martin Raison,
  Alykhan Tejani, Sasank Chilamkurthy, Benoit Steiner, Lu~Fang, Junjie Bai, and
  Soumith Chintala.
\newblock Pytorch: An imperative style, high-performance deep learning library.

\bibitem{pyg}
Matthias Fey and Jan~Eric Lenssen.
\newblock Fast graph representation learning with pytorch geometric.
\newblock volume abs/1903.02428, 2019.

\end{thebibliography}
	\bibliographystyle{unsrt}

\newpage
\DoToC

\newpage
\appendix

\vspace{-0.7em}
\section{Imbalance Formulations}\label{sec:imbdes}
\vspace{-0.7em}
\subsection{Class Distribution Imbalance}
Given the labeled training set $\mathbb{G}_{L} = \{g_i\}_{i=1}^{N_L}$ where each graph $g_i$ belongs to class $y_i \in \mathcal{C}$, we define the class imbalance ratio as:
\begin{equation}
\label{eq:ratioclass}
\rho_{\text{class}} = \frac{\max_{c \in \mathcal{C}} |\{g_i \in \mathbb{G}_{L} | y_i = c\}|}{\min_{c \in \mathcal{C}} |\{g_i \in \mathbb{G}_{L} | y_i = c\}|}
\end{equation}
This formulation captures the disparity between the most and least frequent classes in the training data. In practice, such imbalance leads to biased classifiers that favor majority classes, typically manifesting as low recall rates for minority classes where predictions disproportionately favor the dominant categories.

\subsection{Graph Size Imbalance}
For the complete graph dataset $\mathbb{G} = \{g_i\}_{i=1}^{N}$ where each graph $g_i$ contains $s_i$ nodes, we partition $\mathbb{G}$ into head ($\mathcal{H}_g$) and tail ($\mathcal{T}_g$) sets containing the largest 20\% and remaining 80\% of graphs by size respectively. The size imbalance ratio is:
\begin{equation}
\label{eq:ratiosize}
\rho_{\text{size}} = \frac{\frac{1}{|\mathcal{H}_{\mathrm{g}}|} \sum_{\mathcal{G}_i \in \mathcal{H}_{\mathrm{g}}} s_i}{\frac{1}{|\mathcal{T}_{\mathrm{g}}|} \sum_{\mathcal{G}_j \in \mathcal{T}_{\mathrm{g}}} s_j}
\end{equation}
This structural imbalance creates two distinct challenges: (1) When training and test distributions are aligned, models underperform on tail-size graphs due to insufficient representation; (2) Under distribution shift, models exhibit degraded generalization when extrapolating to size regions sparsely represented in training.

\vspace{-0.7em}
\section{Details of the datsets and Experiments settings}
\vspace{-0.7em}
\subsection{Dataset Descriptions}
\label{app:datasets}
We conducted the experiments on several datasets from the TUDataset series~\cite{tudatasets} and the ogbg-molhiv dataset from OGB Benchmark~\cite{ogbbenchmark}. We use the same train/valid/test dataset splits created by IGL-Bench~\cite{iglbench}. Detailed descriptions of datasets are as follows.

\textbf{PTC-MR}~\cite{ptcmr} is a dataset of 344 organic molecules represented as graphs, labeled for their mutagenic activity on bacteria and their carcinogenic effects on male rats. It is used for chemical compound classification and toxicity prediction, with an unspecified license.

\textbf{D\&D}~\cite{wlstkernel} contains over 1,000 protein structures represented as graphs, where nodes are amino acids and edges connect pairs within 6Å. The label indicates whether a protein is an enzyme. The dataset has an unspecified license.

\textbf{IMDB-BINARY}~\cite{imdbb} is a movie collaboration dataset where nodes represent actors/actresses and edges indicate co-appearances in the same movie. The dataset is licensed under Creative Commons 4.0.

\textbf{REDDIT-BINARY}~\cite{redditb} is a balanced dataset where each graph represents an online discussion thread, with nodes as users and edges indicating reply interactions. The dataset is licensed under Creative Commons 4.0.

\textbf{ogbg-molhiv}~\cite{ogbbenchmark} is a molecular graph classification dataset adopted from MoleculeNet~\cite{wu2018moleculenet} and is among the largest datasets in the collection. Each graph in the dataset represents a molecule, with nodes as atoms and edges as chemical bonds, and input node features are 9-dimensional, including atomic number, chirality, formal charge, and whether the atom is in a ring.

The information of the dataset can be viewed in \cref{tab:datagraph}. The input feature processing of the dataset follows the IGL Bench method. For the datasets PTC\_MR and D\&D, the unique hot code of node label is used. The dimension is the number of node label types, which are 18 and 89, respectively. For IMDB-BINARY and REDDIT-BINARY datasets without node labels, the unique hot code of node degree is used as the input feature, which is 135 and 3062 respectively. The ogbg-molhiv dataset is processed according to the standard method of the OGB Benchmark.

\begin{table}[t]
	\caption{Statistics of benchmark datasets for graph classification.}
	\resizebox{\linewidth}{!}{
		\centering
		\begin{tabular}{lrrrrrrrr}
			\toprule
			\textbf{Dataset} & 
			\textbf{$\#$Graphs} & 
			\textbf{\makecell[r]{Avg.\\ $\#$Nodes}} & 
			\textbf{\makecell[r]{Avg.\\ $\#$Edges}} & 
			\textbf{$\#$Classes} & 
			\textbf{\makecell[r]{$\#$Node \\ Labels}} &
			\textbf{\makecell[r]{$\#$Edge \\ Labels}} &
			\textbf{\makecell[r]{$\#$Node \\ Attrs}} & 
			\textbf{\makecell[r]{$\#$Edge \\ Attrs}} \\
			\midrule
			PTC-MR & 344 & 14.3 & 14.7 & 2 & + & + & - & -~~ \\
			D\&D & 1,178 & 284.3 & 715.7 & 2 & + & - & - & -~~ \\
			IMDB-BINARY & 1,000 & 19.8 & 96.5 & 2 & - & - & - & -~~ \\
			REDDIT-BINARY & 2,000 & 429.6 & 497.8 & 2 & - & - & - & -~~ \\
			ogbg-molhiv & 41,127 & 25.5 & 27.5 & 2 & - & - & +(9) & +(3)~~ \\
			\bottomrule
		\end{tabular}%
	}
	\label{tab:datagraph}%
\end{table}%

\subsection{Experimental Details}
\label{sec:experiment}

\textbf{Computational Environment.} All experiments were conducted using PyTorch~\cite{pytorch} 2.4.0 with CUDA 12.4 and PyTorch-Geometric 2.6.1~\cite{pyg}.

\textbf{Hyperparameter Ranges.} Hyperparameters encompass three parts: (1) general settings, (2) GoG construction hyperparameters, and (3) downstream model hyperparameters. Hyperparameter Ranges are in~\cref{tab:hprange}.

\begin{table}[!h]
	\renewcommand{\cite}[1]{[{\citenum{#1}}]}
	\caption{Hyperparameter search space for the experiments.}
	\resizebox{\linewidth}{!}{
		\centering
		\setlength{\tabcolsep}{1.1em}
		\begin{tabular}{lll}
			\toprule
			\textbf{Parts} & \textbf{Hyperparameter} & \textbf{Search Space} \\
			\midrule
			\multirow{6}[0]{*}{General Settings} & dropout & 0.1, 0.2, 0.5 \\
			& weight decay & 0, $5\times10^{-6}$, $5\times10^{-5}$, $5\times10^{-4}$, $5\times10^{-3}$ \\
			& number of max training epochs & 500 \\
			& learning rate & 0.001,0.005,0.01,0.0125,0.05 \\
			& number of layers & 2, 3, 4 \\
			& hidden size & 32, 64, 128 \\
			& batch size & 128 \\
			\midrule
			\multirow{6}[0]{*}{GoG Construction} & target average node degree $\bar{d}$ & 5, 8, 10 \\
			& minimum node degree $k_{min}$ & 0, 3, 5 \\
			& maximum node degree $k_{min}$ & 100 \\
			& $\rho_1$ in~\cref{rule:rule1} & 5, 9 \\
			& $\rho_2$ in~\cref{rule:rule2} & 1, 3, 9 \\
			& $r$ in~\cref{rule:rule3} & 2, 20, 50  \\
			\midrule
			\multirow{3}[0]{*}{Downstream Tail-GNN~\cite{tailgnn}}
			& number of dropped edges & 3, 5 \\
			& $\mu$ for $\mathcal{L}_m$ & 0.01, 0.001 \\
			& $\eta$ for $\mathcal{L}_d$ & 0.1, 1.0 \\
			\midrule
			\multirow{2}[0]{*}{Downstream GraphSHA~\cite{graphsha}}
			& sampled $\beta$-distribution  & $\beta$(1,100), $\beta$(1,10)\\
			& weighted algorithm & Personalized PageRank, Heat Kernel \\
			\bottomrule
		\end{tabular}%
	}
	\label{tab:hprange}%
\end{table}%
}

\textbf{Edge Homophily Analysis Protocol.} We measure edge homophily ratio $\text{homo}(G) = \frac{\sum_{(i,j)\in E}\mathbb{I}(y_i=y_j)}{|E|}$ for GoG structures, comparing against baseline methods under identical initialization conditions. Evaluation metrics are computed over 10 random seeds to ensure statistical significance.

\vspace{-0.7em}
\section{Proofs of theorems}\label{sec:proofs}
\vspace{-0.7em}
In this section, we provide the proof of the optimization objective modeling of edge homophily; the rationality of the three rules in GoG node degree allocation; and the quality analysis of sampled GoGs for downstream model training.

\subsection{Analysis on Edge Homophily Optimization in Graph-of-Graphs Construction}
\label{sec:appproofhomo}
\begin{proof}
	Proof of~\cref{thm:score2k}.
	
	Consider a GoG graph $G$ with $N$ nodes $V = \{v_1, v_2, \dots, v_N\}$, true labels $Y = \{y_1, \dots, y_N\}$, and $\text{prob}(v_i, y_i) = \sum_{y_j \in Y} \mathbb{I}(y_j = y_i) \cdot \frac{S[i,j]}{\sum_k S[i,k]}$. The graph-of-graphs is constructed such that the average out-degree is constant $K$. The sample size distribution $k = \{k_1, \dots, k_N\}$ satisfies $\sum_i k_i = N\bar{d}$, and the goal is to maximize the expected edge homophily as expressed in \cref{equ:ehomo}:
	\begin{equation}
		\label{equ:apphomo}
		\mathbb{E}(\text{homo}(G)) = \frac{\sum_i k_i \cdot \mathrm{prob}(v_i, y_i)}{N\bar{d}}.
	\end{equation}
	Since the denominator in \cref{equ:apphomo} is constant, the problem reduces to maximizing the numerator $\sum_i k_i \cdot \mathrm{prob}(v_i, y_i)$. This can be viewed as a weight allocation problem.
	Let $t(1), t(2), \dots, t(n)$ denote an arrangement of indices such that:
	\begin{equation}
		\mathrm{prob}(v_{t(1)}, y_{t(1)}) \leq \mathrm{prob}(v_{t(2)}, y_{t(2)}) \leq \dots \leq \mathrm{prob}(v_{t(N)}, y_{t(N)}).
	\end{equation}
	The problem is equivalent to assigning weights $\omega_i = \frac{k_i}{N\bar{d}}$ such that $\sum_i \omega_i = 1$. It is evident that the optimal solution, under no constraints, is:
	\begin{equation}
		\begin{aligned}
			\omega_{t(1)} = \omega_{t(2)} = \dots = \omega_{t(N-1)} = 0, \quad & \omega_{t(N)} = 1, \\
			k_{t(1)} = k_{t(2)} = \dots = k_{t(n-1)} = 0, \quad & k_{t(n)} = N\bar{d}.
		\end{aligned}
	\end{equation}
	However, practical constraints such as non-zero sample sizes and a degree distribution suitable for GNN training impose limits on skewness. Let $k^{opt} = \{k^{opt}_i : 1 \leq i \leq N\}$ denote a sample size sequence satisfying $\sum_i k^{opt}_i = N\bar{d}$ and $k^{opt}_1 \leq k^{opt}_2 \leq \dots \leq k^{opt}_N$. By the rearrangement inequality, for any permutation $\sigma$ of $\{1, 2, \dots, n\}$ and any $\{k_i\}$ satisfying $\sum_1^N k_i = N\bar{d}$:
	\begin{equation}
		\begin{aligned}
			& k^{opt}_1 \cdot \mathrm{prob}(v_{t(1)}) + k^{opt}_2 \cdot \mathrm{prob}(v_{t(2)}) + \dots + k^{opt}_N \cdot \mathrm{prob}(v_{t(N)}) \\
			\geq  & \ k_{\sigma(1)} \cdot \mathrm{prob}(v_{t(1)}) + k_{\sigma(2)} \cdot \mathrm{prob}(v_{t(2)}) + \dots + k_{\sigma(N)} \cdot \mathrm{prob}(v_{t(N)}).
		\end{aligned}
	\end{equation}
	For any two nodes $v_i$ and $v_j$ such that $\mathbb{E}[\mathrm{prob}(v_i, y_i)] > \mathbb{E}[\mathrm{prob}(v_j, y_j)]$ but $k_i < k_j$, exchanging their sample sizes increases edge homophily. Thus, the sample size distribution that maximizes homophily aligns with the $\mathrm{prob}(\cdot)$ ranking of nodes.
\end{proof}

\subsection{Three Rules In Sample Size}
\begin{proof}
\label{proof:r1}
	Proof of~\cref{rule:rule1}[{\bf Label Priority}].

	Consider two graph nodes $v_i$ and $v_j$ in the graph-of-graphs (GoGs), with ground truth labels $y_i = y_j = 0$. Node $v_i$ is labeled, while $v_j$ is unlabeled. Their predicted probabilities are:
	\begin{equation}
		P(\hat{y}_i) = (1, 0), \quad P(\hat{y}_j) = (x, 1-x), \quad x \in (0, 1).
	\end{equation}
	From \cref{eq:similarity}, and $\mathrm{prob}(v_i, y_i) = \sum_{G_j \in \mathcal{G}} \mathbb{I}(y_j = y_i) \cdot \frac{S[i,j]}{\sum_k S[i,k]}$, the $\mathrm{prob}(\cdot)$ value for node $v_i$ is:
	\begin{equation}
	\footnotesize
	\begin{aligned}
	\label{eq:absrule1score}
		& \mathbb{E}(\mathrm{prob}(v_i,0)) = \frac{\mathbb{I}(y_k=0)\sum_k S_{i,k}}{\sum_k S_{i,k}} \\
		& = \frac{(\underset{j,y_j=0}{\sum} P(\hat{y}_k=0))P(\hat{y}_i=0)  + (\underset{k,y_k=0}{\sum}P(\hat{y}_k=1)) P(\hat{y}_i=1)}{(\underset{k,y_k=0}{\sum}P(\hat{y}_k=0) +\underset{k,y_k=1}{\sum}P(\hat{y}_k=0)) P(\hat{y}_i=0) + (\underset{k,y_k=0}{\sum}P(\hat{y}_k=1)+\underset{k,y_k=1}{\sum}P(\hat{y}_k=1)) P(\hat{y}_i=1)}
	\end{aligned}
	\end{equation}
	For ease of expression, we denote four constants: 
	\begin{equation}
	\begin{aligned}
	\gamma_0^0 & = \underset{j,y_k=0}{\sum} P(\hat{y}_k=0), \gamma_0^1 = \underset{k,y_k=0}{\sum} P(\hat{y}_k=1), \\
	\gamma_1^0 & = \underset{j,y_k=1}{\sum} P(\hat{y}_k=0), \gamma_1^1 = \underset{k,y_k=1}{\sum} P(\hat{y}_k=1)
	\end{aligned}
	\end{equation}
	the subscript indicates the true label, and the superscript indicates the predicted label. Therefore, we have,
	\begin{equation}
		\begin{aligned}
				\gamma_0^0 + \gamma_0^1 & = \underset{k,y_k=0}{\sum} 1 \\
				\gamma_1^0 + \gamma_1^1 & = \underset{k,y_k=1}{\sum} 1 \\
			\end{aligned}
	\end{equation}
	For a well trained MLP for binary classification, we assume that the prediction probability of all samples of each class belonging to its own true label is higher than that of the other class, so they satisfy:
	\begin{equation}
		\label{eq:gammaxz}
		\begin{aligned}
		& 0 < \gamma_0^1 < \gamma_0^0 \\ 
		& 0 < \gamma_1^0 < \gamma_1^1.
		\end{aligned}
	\end{equation}
	Let's rewrite the \cref{eq:absrule1score} as a function of $P(\hat{y}_i=0)$,
	\begin{equation}
		\begin{aligned}
		\label{eq:absrule1score2}
		\mathbb{E}(\mathrm{prob}(v_i,0)) = & f_0(P(\hat{y}_i=0)) = \frac{\gamma_0^0 P(\hat{y}_i=0) + \gamma_0^1 (1-P(\hat{y}_i=0))}{(\gamma_0^0+\gamma_1^0) P(\hat{y}_i=0) + (\gamma_0^1+\gamma_1^1) (1-P(\hat{y}_i=0))}, \\
		& f_0(x) = \frac{\gamma_0^0 x + \gamma_0^1 (1-x)}{(\gamma_0^0+\gamma_1^0) x + (\gamma_0^1+\gamma_1^1) (1-x)}
		\end{aligned}
	\end{equation}
	Differentiating $f_0(x)$:
	\begin{equation}
			\label{eq:absrule1score3}
			f_0^{'}(x) = \frac{\gamma_0^0 \gamma_1^1 - \gamma_0^1 \gamma_1^0 }{[(\gamma_0^0+\gamma_1^0) x + (\gamma_0^1+\gamma_1^1) (1-x)]^2} 
	\end{equation}
	According to \cref{eq:gammaxz}, we have,
	\begin{equation}
	\gamma_0^0 \gamma_1^1 - \gamma_0^1 \gamma_1^0 > 0 \Rightarrow f_0^{'}(x) > 0
	\end{equation}
	Since $f_0'(x) > 0$, making $f_0(x)$ an increasing function. Thus, for labeled $v_i$ and unlabeled $v_j$:
	\begin{equation}
	\label{eq:f2Eprob}
	\begin{aligned}
			P(\hat{y}_i=0) = 1 > P(\hat{y}_j=0) & \implies f_0(P(\hat{y}_i=0)) > f_0(P(\hat{y}_j=0)) \\
			& \implies \mathbb{E}[\mathrm{prob}(v_i,0)] > \mathbb{E}[\mathrm{prob}(v_j,0)].
	\end{aligned}
	\end{equation}
	By symmetry, the conclusion holds for class $1$. From \cref{thm:score2k}, a higher score implies $k_i > k_j$ for better homophily.
\end{proof}

\begin{proof}
\label{proof:r2}
	Proof of~\cref{rule:rule2}[{\bf Class Imbalance}].
	
	We analyze and compare the expected scores $\mathbb{E}[\mathrm{prob}(v_i, y_i=0)]$ and $\mathbb{E}[\mathrm{prob}(v_j, y_j=1)]$ in the context of binary classification ($y \in \{0, 1\}$) under three imbalanced training/testing scenarios. The derivation proceeds in six steps.
	
	\textbf{Step 1: Notation and Setup.}

	Let $\mathbb{G}_{\mathrm{L}}$ be the labeled training set, and $\mathbb{G}_{\mathrm{U}}$ the unlabeled part. Denote the class sizes:
	\begin{equation}
	\begin{aligned}
	T_0^L = &  \big| \{g_i \in \mathbb{G}_{\mathrm{L}} : y_i = 0 \} \big|, \quad
	T_1^L = \big| \{g_i \in \mathbb{G}_{\mathrm{L}}: y_i = 1 \} \big|,\\
	T_0^U = &  \big| \{g_j \in \mathbb{G}_{\mathrm{U}} : y_j = 0 \} \big|, \quad
	T_1^U = \big| \{g_j \in \mathbb{G}_{\mathrm{U}} : y_j = 1 \} \big|, \\
	& N_0^U + N_1^U = N^U = |\mathbb{G}_{\mathrm{U}}|, \quad N_0^L + N_1^L = N^L = |\mathbb{G}_{\mathrm{L}}|.
	\end{aligned}
	\end{equation}
	For each test sample $g_a \in \mathbb{G}_{\mathrm{U}}$, $P(\hat{y}_a) = (P(\hat{y}_a=0), P(\hat{y}_a=1))$ denote the predicted probabilities for classes $0$ and $1$ such that $P(\hat{y}_a=0) + P(\hat{y}_a=1) = 1$.
	
	\textbf{Step 2: Precise Definition of $\mathbb{E}[\mathrm{prob}(v_a, y_a)]$.}

	The similarity matrix $S$ is defined as $S = P P^T$, with elements:
	\begin{equation} 
	\begin{aligned}  
		S[a, b] &= P(\hat{y}_a) \cdot P(\hat{y}_b)   
		= \sum_{c \in \{0,1\}} P(\hat{y}_a = c)\, P(\hat{y}_b = c) \\ 
		P(\hat{y}_a) &=   
		\begin{cases}  
			\text{onehot}(y_a), & g_a \in \mathbb{G}_{\mathrm{L}} \\
			\text{Softmax}(\text{logits}(g_a)), & g_a \notin \mathbb{G}_{\mathrm{L}}  
		\end{cases}  
	\end{aligned}
	\end{equation}

	\textbf{Step 3: Random Variables and Expectation.}

	For a training sample $g_i$ and its corresponding GoG node $v_i$ with $y_i = 0$, the score expands to:
	\begin{equation}
	\label{equ:scoredetail}
		\begin{aligned}
	\mathbb{E}[\mathrm{prob}(v_i, y_i=0)] = \frac{T_0^L + X_0^{\text{true}}}{T_0^L + \Omega_0},
	\end{aligned}
	\end{equation}
	where:
	\begin{equation}
		\begin{aligned}
	X_0^{\text{true}} = \sum_{\substack{g_k \in \mathbb{G}_{\mathrm{U}} \\ y_k = 0}} P(\hat{y}_k=0), \quad
	\Omega_0 = \sum_{g_k \in \mathbb{G}_{\mathrm{U}}} P(\hat{y}_k=0).
	\end{aligned}
	\end{equation}
	Similarly, for $g_j$ and its corresponding GoG node $v_j$ with $y_j = 1$,
	\begin{equation}
		\begin{aligned}
	\mathbb{E}[\mathrm{prob}(v_j, y_j=1)] = \frac{T_1^L + X_1^{\text{true}}}{T_1^L + \Omega_1},
	\end{aligned}
	\end{equation}
	where:
	\begin{equation}
		\begin{aligned}
	X_1^{\text{true}} = \sum_{\substack{g_k \in \mathbb{G}_{\mathrm{U}} \\ y_k = 1}} P(\hat{y}_k=1), \quad
	\Omega_1 = \sum_{g_k \in \mathbb{G}_{\mathrm{U}}} P(\hat{y}_k=1).
	\end{aligned}
	\end{equation}
	
	\textbf{Step 4: Asymptotic Convergence and Concentration.}

	Assume the test samples $g_k \in \mathbb{G}_{\mathrm{U}}$ are i.i.d. with class priors $\pi_0 = P(y_k=0)$ and $\pi_1 = P(y_k=1)$, such that $\pi_0 + \pi_1 = 1$. Let $\alpha_0 = \mathbb{E}[P(\hat{y}_k=0) \mid y_k = 0]$ and $\alpha_1 = \mathbb{E}[P(\hat{y}_k=1) \mid y_k = 1]$ denote the expected probabilities assigned to the correct classes. Applying the Law of Large Numbers (LLN):
	\[
	X_0^{\text{true}} \approx \pi_0 \alpha_0 T^U, \quad
	\Omega_0 \approx (\pi_0 \alpha_0 + \pi_1 (1-\alpha_1)) T^U.
	\]
	Substituting these into $\mathbb{E}[\mathrm{prob}(v_i, y_i=0)]$ yields:
	\[
	\mathbb{E}[\mathrm{prob}(v_i, y_i=0)] \approx \frac{T_0^L + \pi_0 \alpha_0 T^U}{T_0^L + (\pi_0 \alpha_0 + \pi_1 (1-\alpha_1)) T^U}.
	\]
	
	\textbf{Step 5: Comparisons Across Scenarios.}

	We evaluate $\mathbb{E}[\mathrm{prob}(v_i, y_i=0)]$ and $\mathbb{E}[\mathrm{prob}(v_j, y_j=1)]$ under three scenarios:
	\begin{itemize}
		\item \textbf{Scenario 1: Class $0$ is dominant in both training and testing.}
		Typically, $\alpha_0 > \alpha_1$ and $\pi_0 > \pi_1$, leading to:
		\[
		\mathbb{E}[\mathrm{prob}(v_i, y_i=0)] > \mathbb{E}[\mathrm{prob}(v_j, y_j=1)].
		\]
		
		\item \textbf{Scenario 2: Balanced training, but class $0$ dominates in testing.}
		Here, $\alpha_0 \approx \alpha_1$ and $\pi_0 > \pi_1$, implying:
		\[
		\mathbb{E}[\mathrm{prob}(v_i, y_i=0)] \geq \mathbb{E}[\mathrm{prob}(v_j, y_j=1)].
		\]
		
		\item \textbf{Scenario 3: Class $0$ is dominant in training, but class $1$ dominates in testing.}
		The reversed class distribution and possible misclassification of truly $1$ samples into $0$ may result in:
		\[
		\mathbb{E}[\mathrm{prob}(v_i, y_i=0)] < \mathbb{E}[\mathrm{prob}(v_j, y_j=1)].
		\]
	\end{itemize}
	
	\textbf{Step 6: Conclusion.}
	These results demonstrate the effect of training/testing imbalance on expected scores. The analysis can be extended to multi-class settings with more intricate assumptions and bounds.
\end{proof}

\begin{proof}
\label{proof:r3}
	Proof of~\cref{rule:rule3} (\textbf{Size Adaptation}).
	
	Let $w_\mathrm{train}=\{s_k \big| g_k \in \mathbb{G}_{L}\}$ denote the size distribution of training graphs and $2r$ be the window width. 
	For each unlabeled node $v_i$ with size $s_i$, define its size window $w_i = [s_i - r, s_i + r]$. 
	
	The \emph{size–generalisation phenomenon} described in the main text
	is captured by the following assumption.
	\begin{assumption}
	\label{as:density}
	For two unlabeled graphs $g_i,g_j$ that share the same ground-truth label $y_i=y_j=c$, $g_i$ has more probability to be classified than $g_j$ due to $|w_\mathrm{train} \cap w_i| > |w_\mathrm{train} \cap w_j|$ according to GNN's size–generalisation phenomenon, therefore $P(\hat{y}_i=c) > P(\hat{y}_j=c)$.
	\end{assumption}

	Following the~\cref{eq:f2Eprob,as:density}, we have 
	\begin{equation}
	\begin{aligned}
	P(\hat{y}_i=c) > P(\hat{y}_j=c) & \implies f_0(P(\hat{y}_i=c)) > f_0(P(\hat{y}_j=c)) \\
	& \implies \mathbb{E}[\mathrm{prob}(v_i,c)] > \mathbb{E}[\mathrm{prob}(v_j,c)].
	\end{aligned}
	\end{equation}
	
\end{proof}

\subsection{Effectiveness of Subgraph Training On Sequence of Sampled GoGs}
\begin{proof}
\label{proof:sampling}
	Effectiveness of GoG Training.
	
	\textbf{I. Problem Setting and Goals}

	Given a directed weighted graph \( G = (V, E) \) with \( N \) nodes and a pairwise graph similarity matrix
	\begin{equation}
		S \in [0,1]^{N \times N},
	\end{equation}
	as weights in importance-sampling, we have:
	\begin{enumerate}
		\item If two labeled nodes \( i \) and \( j \) belong to the same class, then \( S[i,j] = 1 \).
		\item If two labeled nodes \( i \) and \( j \) belong to different classes, then \( S[i,j] = 0 \) (treated as no edge).
		\item Otherwise, each edge weight \( S[i,j] \) lies in the interval \([0,1]\).
	\end{enumerate}
	We employ a Graph Neural Network (GNN), for instance a GCN, for a semi-supervised node classification task on \( G \). Since \( G \) can be almost a complete directed graph (potentially \( O(N^2) \) edges), training a GCN directly on the full graph may be computationally expensive or memory intensive.
	Hence, we introduce a \emph{sampling mechanism}:
	\begin{itemize}
		\item For each node \( i \), we sample \( k_i \) outgoing edges.
		\item The probability of sampling edge \((i,j)\) is proportional to its weight (similarity) \( S[i,j] \).
		\item In the resulting sampled subgraph, each chosen edge is assigned weight 1, while edges not chosen are considered absent.
	\end{itemize}
	After sampling a subgraph in each round, we perform one forward pass and one backward pass (updating the GCN parameters). Repeating this for \( t \) rounds yields the final node embeddings \( H^{(t)} \). Our goal is to prove:
	\begin{itemize}
		\item \textbf{Expectation Approximation:} The expected final embeddings \( H^{(t)} \) approach those obtained by training a weighted GCN on the entire graph \( G \), denoted \( H_{\text{full}} \).
		\item \textbf{Variance Reduction:} As \( t \) grows, the variance of these embeddings decreases, typically at a rate on the order of \( 1/t \).
	\end{itemize}
	
	\textbf{II. Sampling Mechanism and Subgraph Representation}
	
	\textit{(a) Sampling Probabilities.}
	
	In one sampling round, the probability of selecting edge \((i,j)\) is
	\begin{equation}
		p_{i,j} = \frac{k_i \,S[i,j]}{\sum_m S[i,m]}.
	\end{equation}
	We assume \( k_i \) is chosen so that such sampling is feasible.

	\textit{(b) Adjacency Matrix of the Sampled Subgraph.}

	Define the adjacency matrix of the sampled subgraph in the \( s \)-th round by
	\begin{equation}
		A_s \in \{0,1\}^{n\times n},
	\end{equation}
	where
	\begin{equation}
		A_s(i,j) = 
		\begin{cases}
			1, & \text{if edge }(i,j)\text{ is sampled},\\
			0, & \text{otherwise}.
		\end{cases}
	\end{equation}
	Then
	\begin{equation}
		\mathbb{E}[A_s(i,j)] = p_{i,j} =  \frac{k_i \,S[i,j]}{\sum_m S[i,m]}.
	\end{equation}

	\textit{(c) Normalization.} 
 
	GCNs often use a normalized adjacency (e.g.\ \(\hat{A} = D^{-\tfrac12} A \,D^{-\tfrac12}\)). For the sampled subgraph, let
	\begin{equation}
		\hat{A}_s = \mathrm{Norm}(A_s).
	\end{equation}
	Under unbiased sampling assumptions, we typically have
	\begin{equation}
		\mathbb{E}[\hat{A}_s] \approx \hat{W} = \mathrm{Norm}(W).
	\end{equation}
	
	\textbf{III. GCN Training via Randomized Subgraphs (SGD Perspective)}
	
	\textit{(a) Parameterization and Loss Function.}  
	
	Let all GCN parameters be
	\begin{equation}
		\Theta.
	\end{equation}
	We aim to minimize a loss function \(\mathcal{L}(\Theta)\) over the full graph, for instance a cross-entropy for semi-supervised learning. The full gradient is
	\begin{equation}
		\nabla_{\Theta}\,\mathcal{L}_{\text{full}}(\Theta).
	\end{equation}

	\textit{(b) Unbiased Gradient Estimates.}  
	
	When we sample a subgraph \(A_s\) and compute a local loss \(\mathcal{L}_s(\Theta)\), the resulting gradient
	\begin{equation}
		g_s = \nabla_{\Theta}\,\mathcal{L}_s(\Theta)
	\end{equation}
	is designed to be an unbiased estimator of the full gradient:
	\begin{equation}
		\mathbb{E}[g_s] = \nabla_{\Theta}\,\mathcal{L}_{\text{full}}(\Theta).
	\end{equation}

	\textit{(c) Parameter Updates.}

	In the \( s \)-th iteration, given learning rate \(\eta_s\),
	\begin{equation}
		\Theta^{(s)} 
		= 
		\Theta^{(s-1)} - \eta_s \,\nabla_{\Theta}\,\mathcal{L}_s(\Theta^{(s-1)}).
	\end{equation}
	This matches the standard Stochastic Gradient Descent (SGD) framework.
	
	\textbf{IV. Expected Node Embeddings}

	\textit{(a) Relationship Between Embeddings and Parameters.}  
	
	Let
	\begin{equation}
		f(\cdot; \Theta): \mathbb{R}^{N \times d_0} \to \mathbb{R}^{N \times d_L}
	\end{equation}
	denote the GCN forward mapping from input features \( X \) to embeddings. Then
	\begin{equation}
		H(\Theta) = f(X; \Theta).
	\end{equation}
	Suppose \(\Theta_{\text{full}}\) is the optimal parameter (in the full-graph sense), and
	\begin{equation}
		H_{\text{full}} = f\bigl(X; \Theta_{\text{full}}\bigr).
	\end{equation}

	\textit{(b) Convergence in Expectation.}

	Under the unbiased gradient assumption and appropriate SGD theory (e.g.\ Robbins-Monro),
	\begin{equation}
		\Theta^{(s)} 
		\;\to\; 
		\Theta_{\text{full}}
		\quad
		\text{in expectation, as } s \to \infty.
	\end{equation}
	Hence defining
	\begin{equation}
		H^{(t)} = f\bigl(X; \Theta^{(t)}\bigr),
	\end{equation}
	we obtain
	\begin{equation}
		\lim_{t \to \infty} \mathbb{E}[H^{(t)}] = H_{\text{full}},
	\end{equation}
	assuming continuity of \(f\). This proves \(\mathbb{E}[H^{(t)}]\) converges to the full-graph solution.
	
	\textbf{V. Variance Analysis and Rate of Convergence}
	
	\textit{(a) Parameter Variance.}  
	
	If
	\begin{equation}
		g_s = \nabla_{\Theta}\,\mathcal{L}_s(\Theta^{(s-1)})
	\end{equation}
	has variance bounded by \(\Sigma_0\), then the updates
	\begin{equation}
		\Theta^{(s)} = \Theta^{(s-1)} - \eta_s \,g_s
	\end{equation}
	can be controlled by choosing a diminishing learning rate, for example \(\eta_s = \eta_0 / s\). Classical results show that
	\begin{equation}
		\mathrm{Var}(\Theta^{(t)}) = \mathcal{O}\!\Bigl(\frac{1}{t}\Bigr).
	\end{equation}
	\textit{(b) Embedding Variance.}  

	For the node embeddings \( H^{(t)} = f(X; \Theta^{(t)}) \), if we linearize around \(\Theta_{\text{full}}\):
	\begin{equation}
		H^{(t)} 
		\approx
		H_{\text{full}} + J_{\Theta_{\text{full}}}\,\bigl(\Theta^{(t)} - \Theta_{\text{full}}\bigr),
	\end{equation}
	where \(J_{\Theta_{\text{full}}}\) is the Jacobian w.r.t.\ \(\Theta\). Therefore,
	\begin{equation}
		\mathrm{Var}\bigl(H^{(t)}\bigr)
		\approx
		J_{\Theta_{\text{full}}}\,\mathrm{Var}\bigl(\Theta^{(t)}\bigr)\,J_{\Theta_{\text{full}}}^\top
		=
		\mathcal{O}\!\Bigl(\frac{1}{t}\Bigr).
	\end{equation}
	Hence the variance of the embeddings also decays on the order of \( 1/t \).
	
	\textbf{VI. Final Conclusions}

	\textbf{Expectation Consistency:} 

	Every subgraph-based gradient is an unbiased estimator of the full-graph objective’s gradient. Therefore, \(\Theta^{(t)}\) converges in expectation to \(\Theta_{\text{full}}\), and
	\begin{equation}
		\lim_{t \to \infty} \mathbb{E}[H^{(t)}] = H_{\text{full}}.
	\end{equation}

	\textbf{Variance Reduction:}  

	With diminishing learning rates, \(\mathrm{Var}(\Theta^{(t)})\) remains bounded and often scales like \(\mathcal{O}(1/t)\). Consequently,
	\begin{equation}
		\mathrm{Var}\bigl(H^{(t)}\bigr) = \mathcal{O}\!\Bigl(\frac{1}{t}\Bigr).
	\end{equation}
	Thus, the final embeddings increasingly concentrate around the full-graph GCN solution as \( t \) grows.
	\emph{Summary.}
	After \( t \) subgraph samplings and training iterations, the final node embeddings’ expectation converges to the result of training a weighted GCN on the entire graph \( G \). The variance of these embeddings decreases at a rate of \(\mathcal{O}(1/t)\), ensuring they become more stable and closer to the full-graph solution as \( t \) increases.
\end{proof}

\vspace{-0.7em}
\section{Limitations and Broader impacts}\label{sec:lbi}
\vspace{-0.7em}
{\bf Limitations} {While SamGoG focuses on imbalanced graph classification task, the GoG technique is not limited to this specific task and can be extended to a broader range of graph learning applications, such as subgraph matching, link prediction, graph regression, and more.
Additionally, as a specialized paradigm in graph learning, the efficiency of GoG still has room for improvement. Future research could focus on optimizing the framework to enhance its efficiency and applicability to larger and more complex datasets, thereby expanding its potential impact in diverse graph-based learning scenarios.}

 {\bf Broader lmpacts} {SamGoG addresses imbalanced graph classification, offering improved detection accuracy for applications like identifying risk groups in social networks and recognizing protein properties. }
\clearpage

\vspace{-0.7em}
\section{Additional Results on IGL-Bench}
\label{sec:addiexp}
\vspace{-0.7em}
\subsection{Performance of Class Imbalance Classification}
\vspace{-0.7em}
\begin{table}[!ht]
	\centering 
	\caption{\textbf{Accuracy} (\% ± stddev) on datasets with changing class imbalance levels over 10 runs, encoded by GIN / GCN. {\colorbox{firstplace}{\color{red}\bf Red}} and  {\colorbox{secondplace}{Blue}} highlight the best and second-best results respectively.}
	\resizebox{0.8\linewidth}{!}{
	\footnotesize
		\centering
		\setlength{\tabcolsep}{0.8em}
\begin{tabular}{lcccc}
			\toprule
			\textbf{Algorithm} & \textbf{PTC-MR}   & \textbf{D\&D} & \textbf{IMDB-B} & \textbf{REDDIT-B} \\
			\midrule
			\multicolumn{5}{c}{Balance (${\rho_{\text{class}}}=5:5$)}\\
			\midrule
			GIN (bb.)~\cite{xu2018powerful} & 50.43\scalebox{0.75}{±2.69} & 64.04\scalebox{0.75}{±3.79} & 66.05\scalebox{0.75}{±2.57} & 76.66\scalebox{0.75}{±4.80} \\
			G\textsuperscript{2}GNN~\cite{g2gnn2022} & 53.70\scalebox{0.75}{±3.87}  & 66.07\scalebox{0.75}{±2.27} & 61.91\scalebox{0.75}{±3.77} & 72.34\scalebox{0.75}{±2.76} \\
			TopoImb~\cite{topoimb2022} & 50.91\scalebox{0.75}{±2.18} & \cellcolor{secondplace}{66.57\scalebox{0.75}{±3.81}} & \cellcolor{secondplace}{66.28\scalebox{0.75}{±1.85}} & 73.99\scalebox{0.75}{±1.18} \\
			DataDec~\cite{datadec2023} & \cellcolor{secondplace}{54.05\scalebox{0.75}{±4.85}} & 64.46\scalebox{0.75}{±1.88} & 64.09\scalebox{0.75}{±5.75} & \cellcolor{firstplace}{\bf\color{red}79.29\scalebox{0.75}{±8.18}} \\
			ImGKB~\cite{imgkb2023} & 53.48\scalebox{0.75}{±3.50} & 65.45\scalebox{0.75}{±2.88} & 50.16\scalebox{0.75}{±0.34} & 50.24\scalebox{0.75}{±0.29} \\
			SamGoG & \cellcolor{firstplace}{\bf\color{red}59.71\scalebox{0.75}{±2.12}} & \cellcolor{firstplace}{\bf\color{red}72.59\scalebox{0.75}{±1.93}} & \cellcolor{firstplace}{\bf\color{red}69.27\scalebox{0.75}{±1.53}} & \cellcolor{secondplace}{77.74\scalebox{0.75}{±0.50}} \\
\bottomrule
		\end{tabular}%
	}%
	\label{tab:res_graph_cls_gin_acc_all}%
\end{table}%

\begin{table}[!ht]
	\centering 
	\caption{\textbf{Balanced Accuracy} (\% ± stddev) on datasets with changing class imbalance levels over 10 runs, encoded by GIN / GCN. {\colorbox{firstplace}{\color{red}\bf Red}} and  {\colorbox{secondplace}{Blue}} highlight the best and second-best results respectively.}
		\resizebox{0.8\linewidth}{!}{
		\footnotesize
		\centering
		\setlength{\tabcolsep}{0.8em}
		\begin{tabular}{lcccc}
			\toprule
			\textbf{Algorithm} & \textbf{PTC-MR}   & \textbf{D\&D} & \textbf{IMDB-B} & \textbf{REDDIT-B} \\
			\midrule
			\multicolumn{5}{c}{Balance (${\rho_{\text{class}}}=5:5$)}\\
			\midrule
			GIN (bb.)~\cite{xu2018powerful} & 51.40\scalebox{0.75}{±1.64} & 63.37\scalebox{0.75}{±3.26} & 66.05\scalebox{0.75}{±2.57} & 76.66\scalebox{0.75}{±4.80} \\
			G\textsuperscript{2}GNN~\cite{g2gnn2022} & 53.42\scalebox{0.75}{±3.72} & 63.79\scalebox{0.75}{±2.13} & 61.91\scalebox{0.75}{±3.77} & 72.34\scalebox{0.75}{±2.76} \\
			TopoImb~\cite{topoimb2022} & \cellcolor{secondplace}{53.99\scalebox{0.75}{±2.19}} & \cellcolor{secondplace}{67.19\scalebox{0.75}{±2.46}} & \cellcolor{secondplace}{66.28\scalebox{0.75}{±1.85}} & 73.99\scalebox{0.75}{±1.18} \\
			DataDec~\cite{datadec2023} & 53.70\scalebox{0.75}{±4.25} & 62.96\scalebox{0.75}{±1.21} & 64.22\scalebox{0.75}{±5.73} & \cellcolor{firstplace}{\bf\color{red}79.80\scalebox{0.75}{±7.38}} \\
			ImGKB~\cite{imgkb2023} & 53.34\scalebox{0.75}{±3.61} & 62.33\scalebox{0.75}{±4.38} & 49.99\scalebox{0.75}{±0.47} & 50.24\scalebox{0.75}{±0.29} \\
			SamGoG & \cellcolor{firstplace}{\bf\color{red}57.19\scalebox{0.75}{±3.33}} & \cellcolor{firstplace}{\bf\color{red}73.45\scalebox{0.75}{±1.07}} & \cellcolor{firstplace}{\bf\color{red}69.28\scalebox{0.75}{±1.53}} & \cellcolor{secondplace}{77.74\scalebox{0.75}{0.50}} \\
\bottomrule
		\end{tabular}%
	}
	\label{tab:res_graph_cls_gin_bacc_all}%
\end{table}%

\begin{table}[!ht]
	\centering 
	\caption{\textbf{Macro-F1} (\% ± stddev) on datasets with changing class imbalance levels over 10 runs, encoded by GIN / GCN. {\colorbox{firstplace}{\color{red}\bf Red}} and  {\colorbox{secondplace}{Blue}} highlight the best and second-best results respectively.}
	\resizebox{0.8\linewidth}{!}{
	\footnotesize
		\centering
		\setlength{\tabcolsep}{0.8em}
		\begin{tabular}{lcccc}
			\toprule
			\textbf{Algorithm} & \textbf{PTC-MR}   & \textbf{D\&D} & \textbf{IMDB-B} & \textbf{REDDIT-B} \\
			\midrule
			\multicolumn{5}{c}{Balance (${\rho_{\text{class}}}=5:5$)}\\
			\midrule
			GIN (bb.)~\cite{xu2018powerful} & 48.85\scalebox{0.75}{±3.33} & 59.87\scalebox{0.75}{±7.24} & 65.24\scalebox{0.75}{±3.21} & 76.20\scalebox{0.75}{±5.00} \\
			G\textsuperscript{2}GNN~\cite{g2gnn2022} & 52.88\scalebox{0.75}{±3.70} & 63.31\scalebox{0.75}{±2.42} & 60.99\scalebox{0.75}{±3.50} & 71.74\scalebox{0.75}{±3.14} \\
			TopoImb~\cite{topoimb2022} & 48.33\scalebox{0.75}{±3.70}  & \cellcolor{secondplace}{65.72\scalebox{0.75}{±3.45}} & \cellcolor{secondplace}{65.85\scalebox{0.75}{±2.29}} & 73.41\scalebox{0.75}{±0.01}  \\
			DataDec~\cite{datadec2023} & \cellcolor{secondplace}{52.98\scalebox{0.75}{±4.65}} & 62.83\scalebox{0.75}{±1.20} & 62.42\scalebox{0.75}{±8.07} & \cellcolor{firstplace}{\bf\color{red}79.02\scalebox{0.75}{±8.63}} \\
			ImGKB~\cite{imgkb2023} & 50.36\scalebox{0.75}{±5.89} & 61.06\scalebox{0.75}{±7.96} & 35.27\scalebox{0.75}{±4.74} & 33.97\scalebox{0.75}{±0.42} \\
			SamGoG & \cellcolor{firstplace}{\bf\color{red}56.03\scalebox{0.75}{±4.56}} & \cellcolor{firstplace}{\bf\color{red}72.08\scalebox{0.75}{±1.63}} & \cellcolor{firstplace}{\bf\color{red}69.04\scalebox{0.75}{±1.57}} & \cellcolor{secondplace}{77.72\scalebox{0.75}{0.52}} \\
\bottomrule
		\end{tabular}%
	}
	\label{tab:res_graph_cls_gin_mf1_all}%
\end{table}%

\begin{table}[!ht]
	\centering 
	\caption{\textbf{Accuracy} (\% ± stddev) on datasets with changing class imbalance levels over 10 runs, encoded by GIN / GCN. {\colorbox{firstplace}{\color{red}\bf Red}} and  {\colorbox{secondplace}{Blue}} highlight the best and second-best results respectively.}
	\resizebox{0.8\linewidth}{!}{
	\footnotesize
		\centering
		\setlength{\tabcolsep}{0.8em}
		\begin{tabular}{lcccc}
			\toprule
			\textbf{Algorithm} & \textbf{PTC-MR}   & \textbf{D\&D} & \textbf{IMDB-B} & \textbf{REDDIT-B} \\
			\midrule
			\multicolumn{5}{c}{Balance (${\rho_{\text{class}}}=5:5$)}\\
			\midrule
			GCN (bb.)~\cite{kipf2017semisupervised} & 48.62\scalebox{0.75}{±7.12}  & 63.97\scalebox{0.75}{±4.09} & \cellcolor{secondplace}{59.76\scalebox{0.75}{±1.78}} & 66.65\scalebox{0.75}{±1.06} \\
			G\textsuperscript{2}GNN~\cite{g2gnn2022} & 44.35\scalebox{0.75}{±4.32} & 61.47\scalebox{0.75}{±7.70} & 58.35\scalebox{0.75}{±3.47} & 67.02\scalebox{0.75}{±2.55} \\
			TopoImb~\cite{topoimb2022} & 50.61\scalebox{0.75}{±6.39} & 56.13\scalebox{0.75}{±4.61} & 49.52\scalebox{0.75}{±1.01} & 58.62\scalebox{0.75}{±6.52} \\
			DataDec~\cite{datadec2023} & \cellcolor{secondplace}{54.23\scalebox{0.75}{±4.16}}  & 63.10\scalebox{0.75}{±2.62} & 58.20\scalebox{0.75}{±4.05} & \cellcolor{secondplace}{69.08\scalebox{0.75}{±4.49}} \\
			ImGKB~\cite{imgkb2023} & 52.83\scalebox{0.75}{±5.45}& \cellcolor{secondplace}{64.93\scalebox{0.75}{±4.44}} & 50.15\scalebox{0.75}{±0.39} & 50.29\scalebox{0.75}{±0.19} \\
			SamGoG & \cellcolor{firstplace}{\bf\color{red}59.49\scalebox{0.75}{±2.68}} & \cellcolor{firstplace}{\bf\color{red}75.13\scalebox{0.75}{±0.90}} & \cellcolor{firstplace}{\bf\color{red}71.70\scalebox{0.75}{±0.79}} & \cellcolor{firstplace}{\bf\color{red}80.05\scalebox{0.75}{±0.23}} \\
\bottomrule
		\end{tabular}%
	}
	\label{tab:res_graph_cls_gcn_acc_all}%
\end{table}%

\begin{table}[!ht]
	\centering 
	\caption{\textbf{Balanced Accuracy} (\% ± stddev) on datasets with changing class imbalance levels over 10 runs, encoded by GIN / GCN. {\colorbox{firstplace}{\color{red}\bf Red}} and  {\colorbox{secondplace}{Blue}} highlight the best and second-best results respectively.}
	\resizebox{0.8\linewidth}{!}{
	\footnotesize
		\centering
		\setlength{\tabcolsep}{0.8em}
		\begin{tabular}{lcccc}
			\toprule
			\textbf{Algorithm} & \textbf{PTC-MR}   & \textbf{D\&D} & \textbf{IMDB-B} & \textbf{REDDIT-B} \\
			\midrule
			\multicolumn{5}{c}{Balance (${\rho_{\text{class}}}=5:5$)}\\
			\midrule
			GCN (bb.)~\cite{kipf2017semisupervised} & 50.07\scalebox{0.75}{±0.23} & 62.63\scalebox{0.75}{±2.06} & \cellcolor{secondplace}{59.76\scalebox{0.75}{±1.78}} & 66.65\scalebox{0.75}{±1.06} \\
			G\textsuperscript{2}GNN~\cite{g2gnn2022} & 50.02\scalebox{0.75}{±0.06} & 60.38\scalebox{0.75}{±5.83} & 58.35\scalebox{0.75}{±3.47} & 67.02\scalebox{0.75}{±2.55} \\
			TopoImb~\cite{topoimb2022} & 53.71\scalebox{0.75}{±2.88} & 60.56\scalebox{0.75}{±2.30} & 49.71\scalebox{0.75}{±0.34} & 58.68\scalebox{0.75}{±6.48} \\
			DataDec~\cite{datadec2023} & \cellcolor{secondplace}{54.36\scalebox{0.75}{±3.61}} & 61.41\scalebox{0.75}{±2.42} & 58.11\scalebox{0.75}{±3.96} & \cellcolor{secondplace}{69.08\scalebox{0.75}{±4.38}} \\
			ImGKB~\cite{imgkb2023} & 53.03\scalebox{0.75}{±3.56}  & \cellcolor{secondplace}{62.84\scalebox{0.75}{±2.56}} & 50.15\scalebox{0.75}{±0.39} & 50.29\scalebox{0.75}{±0.19} \\
			SamGoG & \cellcolor{firstplace}{\bf\color{red}56.98\scalebox{0.75}{±2.08}} & \cellcolor{firstplace}{\bf\color{red}73.57\scalebox{0.75}{±0.58}} & \cellcolor{firstplace}{\bf\color{red}71.70\scalebox{0.75}{±0.79}} & \cellcolor{firstplace}{\bf\color{red}80.05\scalebox{0.75}{±0.23}} \\
\bottomrule
		\end{tabular}%
	}
	\label{tab:res_graph_cls_gcn_bacc_all}%
\end{table}%

\begin{table}[!ht]
	\centering 
	\caption{\textbf{Macro-F1} (\% ± stddev) on datasets with changing class imbalance levels over 10 runs, encoded by GIN / GCN. {\colorbox{firstplace}{\color{red}\bf Red}} and  {\colorbox{secondplace}{Blue}} highlight the best and second-best results respectively.}
	\resizebox{0.8\linewidth}{!}{
	\footnotesize
		\centering
		\setlength{\tabcolsep}{0.8em}
		\begin{tabular}{lcccc}
			\toprule
			\textbf{Algorithm} & \textbf{PTC-MR}   & \textbf{D\&D} & \textbf{IMDB-B} & \textbf{REDDIT-B} \\
			\midrule
			\multicolumn{5}{c}{Balance (${\rho_{\text{class}}}=5:5$)}\\
			\midrule
			GCN (bb.)~\cite{kipf2017semisupervised} & 32.82\scalebox{0.75}{±3.29}  & 62.02\scalebox{0.75}{±2.84} & \cellcolor{secondplace}{59.32\scalebox{0.75}{±2.48}} & \cellcolor{secondplace}{67.58\scalebox{0.75}{±1.06}}  \\
			G\textsuperscript{2}GNN~\cite{g2gnn2022} & 31.29\scalebox{0.75}{±2.68}  & 56.99\scalebox{0.75}{±10.85} & 56.41\scalebox{0.75}{±7.64} & 65.63\scalebox{0.75}{±3.26}  \\
			TopoImb~\cite{topoimb2022} & 45.01\scalebox{0.75}{±8.36}  & 55.05\scalebox{0.75}{±4.87} & 33.04\scalebox{0.75}{±0.43} & 50.17\scalebox{0.75}{±12.11}  \\
			DataDec~\cite{datadec2023} & \cellcolor{secondplace}{53.24\scalebox{0.75}{±3.23}}  & 61.35\scalebox{0.75}{±2.55} & 54.26\scalebox{0.75}{±7.60} & 67.37\scalebox{0.75}{±6.20} \\
			ImGKB~\cite{imgkb2023} & 48.53\scalebox{0.75}{±8.31} & \cellcolor{secondplace}{62.34\scalebox{0.75}{±3.64}} & 33.84\scalebox{0.75}{±0.60} & 33.97\scalebox{0.75}{±0.41} \\
			SamGoG & \cellcolor{firstplace}{\bf\color{red}55.89\scalebox{0.75}{±3.20}} & \cellcolor{firstplace}{\bf\color{red}73.72\scalebox{0.75}{±0.68}} & \cellcolor{firstplace}{\bf\color{red}71.69\scalebox{0.75}{±0.79}} & \cellcolor{firstplace}{\bf\color{red}79.95\scalebox{0.75}{±0.19}} \\
\bottomrule
		\end{tabular}%
	}
	\label{tab:res_graph_cls_gcn_mf1_all}%
\end{table}%

\clearpage

\subsection{Performance of Graph Size Imbalance Classification}
\vspace{-0.7em}
\begin{table}[!ht]
	\centering 
	\caption{\textbf{Accuracy} (\% ± stddev) on datasets with changing size imbalance levels over 10 runs, encoded by GIN / GCN. {\colorbox{firstplace}{\color{red}\bf Red}} and  {\colorbox{secondplace}{Blue}} highlight the best and second-best results respectively.}
	\resizebox{0.8\linewidth}{!}{
	\footnotesize
		\centering
		\setlength{\tabcolsep}{0.8em}
		\begin{tabular}{lcccc}
			\toprule
			\textbf{Algorithm} & \textbf{PTC-MR}   & \textbf{D\&D} & \textbf{IMDB-B} & \textbf{REDDIT-B} \\
			\midrule
			Low level $\rho_{\text{size}}$ & 1.60 & 1.56 & 1.64 & 2.79 \\
			\midrule
			GIN (bb.)~\cite{xu2018powerful} & \cellcolor{secondplace}{52.17\scalebox{0.75}{±4.36}} & 65.01\scalebox{0.75}{±0.69} & \cellcolor{secondplace}{66.38\scalebox{0.75}{±4.27}} & 67.41\scalebox{0.75}{±2.32}\\
			SOLT-GNN~\cite{soltgnn2022} & 47.54\scalebox{0.75}{±4.33} & 63.78\scalebox{0.75}{±1.06} & 64.20\scalebox{0.75}{±5.46} & 49.57\scalebox{0.75}{±6.78}\\
			TopoImb~\cite{topoimb2022} & 49.71\scalebox{0.75}{±1.98}  & \cellcolor{secondplace}{65.16\scalebox{0.75}{±1.76}} & 66.00\scalebox{0.75}{±2.41} & \cellcolor{secondplace}{69.47\scalebox{0.75}{±3.70}}\\
			SamGoG & \cellcolor{firstplace}{\bf\color{red}53.86\scalebox{0.75}{±2.12}} & \cellcolor{firstplace}{\bf\color{red}68.22\scalebox{0.75}{±2.40}} & \cellcolor{firstplace}{\bf\color{red}66.82\scalebox{0.75}{±0.85}} & \cellcolor{firstplace}{\bf\color{red}72.20\scalebox{0.75}{±4.24}} \\
			\midrule    
			Mid level $\rho_{\text{size}}$ & 2.90  & 2.53 & 2.11 & 4.92 \\
			\midrule
			GIN (bb.)~\cite{xu2018powerful} & 51.38\scalebox{0.75}{±6.78} & 61.46\scalebox{0.75}{±2.43} & 65.08\scalebox{0.75}{±5.78} & 68.32\scalebox{0.75}{±1.77} \\
\bottomrule
		\end{tabular}%
	}
	\label{tab:res_graph_topo_gin_acc_all}%
\end{table}%

\begin{table}[!ht]
	\centering 
	\caption{\textbf{Balanced Accuracy} (\% ± stddev) on datasets with changing size imbalance levels over 10 runs, encoded by GIN / GCN. {\colorbox{firstplace}{\color{red}\bf Red}} and  {\colorbox{secondplace}{Blue}} highlight the best and second-best results respectively.}
	\resizebox{0.8\linewidth}{!}{
	\footnotesize
		\centering
		\setlength{\tabcolsep}{0.8em}
		\begin{tabular}{lcccc}
			\toprule
			\textbf{Algorithm} & \textbf{PTC-MR}   & \textbf{D\&D} & \textbf{IMDB-B} & \textbf{REDDIT-B} \\
			\midrule
			Low level $\rho_{\text{size}}$ & 1.60 & 1.56 & 1.64 & 2.79 \\
			\midrule
			GIN (bb.)~\cite{xu2018powerful} & \cellcolor{secondplace}{51.08\scalebox{0.75}{±2.07}}  & 63.46\scalebox{0.75}{±0.92} & \cellcolor{secondplace}{66.38\scalebox{0.75}{±4.27}} & 67.41\scalebox{0.75}{±2.32} \\
			SOLT-GNN~\cite{soltgnn2022} & 45.47\scalebox{0.75}{±2.10} & 62.58\scalebox{0.75}{±0.61} & 64.20\scalebox{0.75}{±5.46} & 49.57\scalebox{0.75}{±6.78}  \\
			TopoImb~\cite{topoimb2022} & 50.07\scalebox{0.75}{±2.77} & \cellcolor{secondplace}{67.24\scalebox{0.75}{±1.53}} & 66.00\scalebox{0.75}{±2.41} & \cellcolor{secondplace}{69.47\scalebox{0.75}{±3.70}} \\
			SamGoG & \cellcolor{firstplace}{\bf\color{red}52.66\scalebox{0.75}{±2.33}} & \cellcolor{firstplace}{\bf\color{red}71.20\scalebox{0.75}{±1.78}} & \cellcolor{firstplace}{\bf\color{red}66.82\scalebox{0.75}{±0.85}} & \cellcolor{firstplace}{\bf\color{red}72.20\scalebox{0.75}{±4.24}} \\
			\midrule    
			Mid level $\rho_{\text{size}}$ & 2.90  & 2.53 & 2.11 & 4.92 \\
			\midrule
			GIN (bb.)~\cite{xu2018powerful} & 49.83\scalebox{0.75}{±1.30} & 62.22\scalebox{0.75}{±0.98} & 65.08\scalebox{0.75}{±5.78} & 68.32\scalebox{0.75}{±1.77}  \\
\bottomrule
		\end{tabular}%
	}
	\label{tab:res_graph_topo_gin_bacc_all}%
\end{table}%

\begin{table}[!ht]
	\centering 
	\caption{\textbf{Macro-F1} (\% ± stddev) on datasets with changing size imbalance levels over 10 runs, encoded by GIN / GCN. {\colorbox{firstplace}{\color{red}\bf Red}} and  {\colorbox{secondplace}{Blue}} highlight the best and second-best results respectively.}
	\resizebox{0.8\linewidth}{!}{
	\footnotesize
		\centering
		\setlength{\tabcolsep}{0.8em}
		\begin{tabular}{lcccc}
			\toprule
			\textbf{Algorithm} & \textbf{PTC-MR}   & \textbf{D\&D} & \textbf{IMDB-B} & \textbf{REDDIT-B} \\
			\midrule
			Low level $\rho_{\text{size}}$ & 1.60 & 1.56 & 1.64 & 2.79 \\
			\midrule
			GIN (bb.)~\cite{xu2018powerful} & 48.26\scalebox{0.75}{±3.11} & 63.39\scalebox{0.75}{±0.84} & 65.67\scalebox{0.75}{±5.40} & 66.14\scalebox{0.75}{±1.90}  \\
			SOLT-GNN~\cite{soltgnn2022} & 43.63\scalebox{0.75}{±1.91}  & 62.34\scalebox{0.75}{±0.66} & 63.66\scalebox{0.75}{±5.94} & 49.17\scalebox{0.75}{±6.76}  \\
			TopoImb~\cite{topoimb2022} & \cellcolor{secondplace}{48.87\scalebox{0.75}{±2.90}} & \cellcolor{secondplace}{64.97\scalebox{0.75}{±1.65}} & \cellcolor{secondplace}{65.71\scalebox{0.75}{±2.62}} & \cellcolor{secondplace}{69.15\scalebox{0.75}{±3.83}} \\
			SamGoG & \cellcolor{firstplace}{\bf\color{red}52.52\scalebox{0.75}{±2.28}} & \cellcolor{firstplace}{\bf\color{red}68.11\scalebox{0.75}{±2.41}} & \cellcolor{firstplace}{\bf\color{red}66.20\scalebox{0.75}{±1.16}} & \cellcolor{firstplace}{\bf\color{red}70.50\scalebox{0.75}{±5.03}} \\
			\midrule    
			Mid level $\rho_{\text{size}}$ & 2.90  & 2.53 & 2.11 & 4.92 \\
			\midrule
			GIN (bb.)~\cite{xu2018powerful} & 38.58\scalebox{0.75}{±5.97} & 60.89\scalebox{0.75}{±1.84} & 63.28\scalebox{0.75}{±9.06} & 67.20\scalebox{0.75}{±2.32}  \\
\bottomrule
		\end{tabular}%
	}
	\label{tab:res_graph_topo_gin_mf1_all}%
\end{table}%

\begin{table}[!ht]
	\centering 
	\caption{\textbf{Accuracy} (\% ± stddev) on datasets with changing size imbalance levels over 10 runs, encoded by GIN / GCN. {\colorbox{firstplace}{\color{red}\bf Red}} and  {\colorbox{secondplace}{Blue}} highlight the best and second-best results respectively.}
	\resizebox{0.8\linewidth}{!}{
	\footnotesize
		\centering
		\setlength{\tabcolsep}{0.8em}
		\begin{tabular}{lcccc}
			\toprule
			\textbf{Algorithm} & \textbf{PTC-MR}   & \textbf{D\&D} & \textbf{IMDB-B} & \textbf{REDDIT-B} \\
			\midrule
			Low level $\rho_{\text{size}}$ & 1.60 & 1.56 & 1.64 & 2.79 \\
			\midrule
			GCN (bb.)~\cite{kipf2017semisupervised} & \cellcolor{secondplace}{51.59\scalebox{0.75}{±7.07}} & 64.82\scalebox{0.75}{±0.75} & \cellcolor{secondplace}{69.15\scalebox{0.75}{±1.44}} & 53.44\scalebox{0.75}{±1.78} \\
			SOLT-GNN~\cite{soltgnn2022} & 51.45\scalebox{0.75}{±3.13} & \cellcolor{secondplace}{65.69\scalebox{0.75}{±2.36}} & 68.23\scalebox{0.75}{±2.48} & 37.58\scalebox{0.75}{±3.04} \\
			TopoImb~\cite{topoimb2022} & 46.74\scalebox{0.75}{±1.57} & 56.18\scalebox{0.75}{±2.10} & 68.20\scalebox{0.75}{±0.70} & \cellcolor{secondplace}{56.94\scalebox{0.75}{±3.23}}  \\
			SamGoG & \cellcolor{firstplace}{\bf\color{red}52.83\scalebox{0.75}{±0.90}} & \cellcolor{firstplace}{\bf\color{red}71.06\scalebox{0.75}{±3.54}} & \cellcolor{firstplace}{\bf\color{red}70.53\scalebox{0.75}{±0.87}} & \cellcolor{firstplace}{\bf\color{red}72.60\scalebox{0.75}{±1.72}} \\
			\midrule    
			Mid level $\rho_{\text{size}}$ & 2.90  & 2.53 & 2.11 & 4.92 \\
			\midrule
			GCN (bb.)~\cite{kipf2017semisupervised} & 50.00\scalebox{0.75}{±5.87}  & 63.99\scalebox{0.75}{±2.16} & 67.95\scalebox{0.75}{±2.82} & 67.96\scalebox{0.75}{±0.89} \\
\bottomrule
		\end{tabular}%
	}
	\label{tab:res_graph_topo_gcn_acc_all}%
\end{table}%

\begin{table}[!ht]
	\centering 
	\caption{\textbf{Balanced Accuracy} (\% ± stddev) on datasets with changing size imbalance levels over 10 runs, encoded by GIN / GCN. {\colorbox{firstplace}{\color{red}\bf Red}} and  {\colorbox{secondplace}{Blue}} highlight the best and second-best results respectively.}
	\resizebox{0.8\linewidth}{!}{
	\footnotesize
		\centering
		\setlength{\tabcolsep}{0.8em}
		\begin{tabular}{lcccc}
			\toprule
			\textbf{Algorithm} & \textbf{PTC-MR}   & \textbf{D\&D} & \textbf{IMDB-B} & \textbf{REDDIT-B} \\
			\midrule
			Low level $\rho_{\text{size}}$ & 1.60 & 1.56 & 1.64 & 2.79 \\
			\midrule
			GCN (bb.)~\cite{kipf2017semisupervised} & \cellcolor{secondplace}{50.47\scalebox{0.75}{±0.75}} & 63.13\scalebox{0.75}{±0.43} & \cellcolor{secondplace}{69.15\scalebox{0.75}{±1.44}} & 53.44\scalebox{0.75}{±1.78} \\
			SOLT-GNN~\cite{soltgnn2022} & 48.24\scalebox{0.75}{±1.20} & \cellcolor{secondplace}{63.81\scalebox{0.75}{±1.86}} & 68.23\scalebox{0.75}{±2.48} & 37.58\scalebox{0.75}{±3.04} \\
			TopoImb~\cite{topoimb2022} & 50.43\scalebox{0.75}{±0.87} & 57.64\scalebox{0.75}{±0.94} & 68.20\scalebox{0.75}{±0.70} & \cellcolor{secondplace}{56.94\scalebox{0.75}{±3.23}} \\
			SamGoG & \cellcolor{firstplace}{\bf\color{red}53.33\scalebox{0.75}{±0.27}}  & \cellcolor{firstplace}{\bf\color{red}70.46\scalebox{0.75}{±3.84}} & \cellcolor{firstplace}{\bf\color{red}70.52\scalebox{0.75}{±0.87}} & \cellcolor{firstplace}{\bf\color{red}72.60\scalebox{0.75}{±1.72}} \\
			\midrule    
			Mid level $\rho_{\text{size}}$ & 2.90  & 2.53 & 2.11 & 4.92 \\
			\midrule
			GCN (bb.)~\cite{kipf2017semisupervised} & 49.74\scalebox{0.75}{±2.52} & 58.27\scalebox{0.75}{±4.65} & 67.95\scalebox{0.75}{±2.82} & 67.96\scalebox{0.75}{±0.89} \\
\bottomrule
		\end{tabular}%
	}
	\label{tab:res_graph_topo_gcn_bacc_all}%
\end{table}%

\begin{table}[!ht]
	\centering 
	\caption{\textbf{Macro-F1} (\% ± stddev) on datasets with changing size imbalance levels over 10 runs, encoded by GIN / GCN. {\colorbox{firstplace}{\color{red}\bf Red}} and  {\colorbox{secondplace}{Blue}} highlight the best and second-best results respectively.}
	\resizebox{0.8\linewidth}{!}{
	\footnotesize
		\centering
		\setlength{\tabcolsep}{0.8em}
		\begin{tabular}{lcccc}
			\toprule
			\textbf{Algorithm} & \textbf{PTC-MR}   & \textbf{D\&D} & \textbf{IMDB-B} & \textbf{REDDIT-B} \\
			\midrule
			Low level $\rho_{\text{size}}$ & 1.60 & 1.56 & 1.64 & 2.79 \\
			\midrule
			GCN (bb.)~\cite{kipf2017semisupervised} & 36.72\scalebox{0.75}{±5.58} & 63.09\scalebox{0.75}{±0.46} & 69.02\scalebox{0.75}{±1.52} & 49.68\scalebox{0.75}{±8.19} \\
			SOLT-GNN~\cite{soltgnn2022} & \cellcolor{secondplace}{45.00\scalebox{0.75}{±3.14}} & \cellcolor{secondplace}{63.82\scalebox{0.75}{±2.01}} & \cellcolor{secondplace}{67.88\scalebox{0.75}{±2.35}} & 35.00\scalebox{0.75}{±2.77} \\
			TopoImb~\cite{topoimb2022} & 44.72\scalebox{0.75}{±2.61} & 55.77\scalebox{0.75}{±1.91} & 67.78\scalebox{0.75}{±1.00} & \cellcolor{secondplace}{56.09\scalebox{0.75}{±3.68}} \\
			SamGoG & \cellcolor{firstplace}{\bf\color{red}52.51\scalebox{0.75}{±0.51}} & \cellcolor{firstplace}{\bf\color{red}70.02\scalebox{0.75}{±3.71}} & \cellcolor{firstplace}{\bf\color{red}70.45\scalebox{0.75}{±0.84}} & \cellcolor{firstplace}{\bf\color{red}71.77\scalebox{0.75}{±2.05}} \\
			\midrule    
			Mid level $\rho_{\text{size}}$ & 2.90  & 2.53 & 2.11 & 4.92 \\
			\midrule
			GCN (bb.)~\cite{kipf2017semisupervised} & 40.94\scalebox{0.75}{±5.67}  & 54.22\scalebox{0.75}{±8.96} & 67.80\scalebox{0.75}{±2.84} & 67.18\scalebox{0.75}{±1.39} \\
\bottomrule
		\end{tabular}%
	}
	\label{tab:res_graph_topo_gcn_mf1_all}%
\end{table}%

\clearpage

\subsection{Performance on Different Downstream Models}
\label{sec:graphsha}
\vspace{-0.7em}
\begin{table}[!ht]
\centering 
\caption{\textbf{Accuracy} score (\% ± stddev) on \textbf{D\&D} dataset with different class imbalance levels over 10 runs. {\colorbox{firstplace}{\color{red}\bf Red}} and  {\colorbox{secondplace}{Blue}} highlight the best and second-best results respectively.}
\resizebox{0.8\linewidth}{!}{
\footnotesize
\centering
\setlength{\tabcolsep}{0.8em}
\begin{tabular}{lccc}
\toprule
\textbf{Algorithm} & \textbf{Balance}& \textbf{Low Imbalance} & \textbf{High Imbalance} \\
\midrule
$\rho_{\text{class}}$ & 1:1 & 7:3 & 9:1 \\
\midrule
GIN (bb.)~\cite{xu2018powerful} & 64.04\scalebox{0.75}{±3.79} & 51.05\scalebox{0.75}{±5.07} & 41.54\scalebox{0.75}{±6.57} \\
G\textsuperscript{2}GNN~\cite{g2gnn2022} & 66.07\scalebox{0.75}{±2.27} & 56.29\scalebox{0.75}{±7.30} & 55.38\scalebox{0.75}{±15.60} \\
TopoImb~\cite{topoimb2022} & 66.57\scalebox{0.75}{±3.81} & 49.97\scalebox{0.75}{±7.24} & 39.12\scalebox{0.75}{±1.62} \\
DataDec~\cite{datadec2023} & 64.46\scalebox{0.75}{±1.88} & 63.51\scalebox{0.75}{±1.62} & 65.77\scalebox{0.75}{±2.71} \\
ImGKB~\cite{imgkb2023} & 65.45\scalebox{0.75}{±2.88} & 65.85\scalebox{0.75}{±3.70} & 59.99\scalebox{0.75}{±7.57} \\
\midrule
SamGoG (w/ TailGNN~\cite{tailgnn}) 
& \cellcolor{firstplace}{\bf\color{red}72.59\scalebox{0.75}{±1.93}} 
& \cellcolor{secondplace}{73.31}\scalebox{0.75}{±1.99} 
& \cellcolor{firstplace}{\bf\color{red}71.94\scalebox{0.75}{±1.73}} \\
SamGoG (w/ GraphSHA~\cite{graphsha}) 
& \cellcolor{secondplace}{71.73}\scalebox{0.75}{±2.53} 
& \cellcolor{firstplace}{\bf\color{red}73.99\scalebox{0.75}{±0.98}} 
& \cellcolor{secondplace}{70.35}\scalebox{0.75}{±1.62} \\
\bottomrule
\end{tabular}
}
\label{tab:res_dd_cls_gin_acc}%
\end{table}

\begin{table}[!ht]
\centering 
\caption{\textbf{Accuracy} score (\% ± stddev) on \textbf{D\&D} dataset with different class imbalance levels over 10 runs. {\colorbox{firstplace}{\color{red}\bf Red}} and  {\colorbox{secondplace}{Blue}} highlight the best and second-best results respectively.}
\resizebox{0.8\linewidth}{!}{
\footnotesize
\centering
\setlength{\tabcolsep}{0.8em}
\begin{tabular}{lccc}
\toprule
\textbf{Algorithm} & \textbf{Balance}& \textbf{Low Imbalance} & \textbf{High Imbalance} \\
\midrule
$\rho_{\text{class}}$ & 1:1 & 7:3 & 9:1 \\
\midrule
GCN (bb.)~\cite{kipf2017semisupervised} & 63.97\scalebox{0.75}{±4.09} & 58.32\scalebox{0.75}{±1.51} & 42.48\scalebox{0.75}{±2.61} \\
G\textsuperscript{2}GNN~\cite{g2gnn2022} & 61.47\scalebox{0.75}{±7.70} & 65.65\scalebox{0.75}{±5.41} & 44.59\scalebox{0.75}{±4.72} \\
TopoImb~\cite{topoimb2022} & 56.13\scalebox{0.75}{±4.61} & 44.63\scalebox{0.75}{±4.82} & 36.50\scalebox{0.75}{±1.44} \\
DataDec~\cite{datadec2023} & 63.10\scalebox{0.75}{±2.62} & 64.66\scalebox{0.75}{±2.04} & 65.07\scalebox{0.75}{±2.40} \\
ImGKB~\cite{imgkb2023} & 64.93\scalebox{0.75}{±4.44} & 67.44\scalebox{0.75}{±3.50} & 58.21\scalebox{0.75}{±10.60} \\
\midrule
SamGoG (w/ TailGNN~\cite{tailgnn}) &
\cellcolor{firstplace}{\bf\color{red}75.13\scalebox{0.75}{±0.90}} & 
\cellcolor{secondplace}{74.81}\scalebox{0.75}{±0.99} & 
\cellcolor{secondplace}{66.90}\scalebox{0.75}{±1.83} \\
SamGoG (w/ GraphSHA~\cite{graphsha}) &
\cellcolor{secondplace}{73.91}\scalebox{0.75}{±1.65} &
\cellcolor{firstplace}{\bf\color{red}74.94\scalebox{0.75}{±1.85}} &
\cellcolor{firstplace}{\bf\color{red}70.07\scalebox{0.75}{±2.94}} \\
\bottomrule
\end{tabular}
}
\label{tab:res_dd_cls_gcn_acc}%
\end{table}

\begin{table}[!ht]
\centering 
\caption{\textbf{Accuracy} score (\% ± stddev) on \textbf{D\&D} dataset with different graph size imbalance levels over 10 runs. {\colorbox{firstplace}{\color{red}\bf Red}} and  {\colorbox{secondplace}{Blue}} highlight the best and second-best results respectively.}
\resizebox{0.8\linewidth}{!}{
\footnotesize
\centering
\setlength{\tabcolsep}{0.8em}
\begin{tabular}{lccc}
\toprule
\textbf{Algorithm} & \textbf{Balance}& \textbf{Low Imbalance} & \textbf{High Imbalance} \\
\midrule
$\rho_{\text{size}}$ & 1.56 & 2.53 & 8,91 \\
\midrule
GIN (bb.)~\cite{xu2018powerful} & 
65.01\scalebox{0.75}{±0.69} & 
61.46\scalebox{0.75}{±2.43} & 
60.68\scalebox{0.75}{±6.89} \\

SOLT-GNN~\cite{soltgnn2022} & 
63.78\scalebox{0.75}{±1.06} & 
63.33\scalebox{0.75}{±1.86} & 
64.97\scalebox{0.75}{±3.24} \\

TopoImb~\cite{topoimb2022} & 
65.16\scalebox{0.75}{±1.76} & 
65.99\scalebox{0.75}{±1.25} & 
64.16\scalebox{0.75}{±2.96} \\

\midrule
SamGoG (w/ TailGNN~\cite{tailgnn}) &
\cellcolor{firstplace}{\bf\color{red}68.22\scalebox{0.75}{±2.40}} & 
\cellcolor{firstplace}{\bf\color{red}71.91\scalebox{0.75}{±1.22}} & 
\cellcolor{secondplace}{69.15}\scalebox{0.75}{±1.87} \\
SamGoG (w/ GraphSHA~\cite{graphsha}) &
\cellcolor{secondplace}{66.71}\scalebox{0.75}{±2.07} & 
\cellcolor{secondplace}{71.17}\scalebox{0.75}{±1.53} & 
\cellcolor{firstplace}{\bf\color{red}70.00\scalebox{0.75}{±1.83}} \\
\bottomrule
\end{tabular}
}
\label{tab:res_dd_size_gin_acc}%
\end{table}

\begin{table}[!ht]
\centering 
\caption{\textbf{Accuracy} score (\% ± stddev) on \textbf{D\&D} dataset with different graph size imbalance levels over 10 runs. {\colorbox{firstplace}{\color{red}\bf Red}} and  {\colorbox{secondplace}{Blue}} highlight the best and second-best results respectively.}
\resizebox{0.8\linewidth}{!}{
\footnotesize
\centering
\setlength{\tabcolsep}{0.8em}
\begin{tabular}{lccc}
\toprule
\textbf{Algorithm} & \textbf{Balance}& \textbf{Low Imbalance} & \textbf{High Imbalance} \\
\midrule
$\rho_{\text{size}}$ & 1.56 & 2.53 & 8,91 \\
\midrule
GCN (bb.)~\cite{kipf2017semisupervised} & 
64.82\scalebox{0.75}{±0.75} & 
63.99\scalebox{0.75}{±2.16} & 
60.98\scalebox{0.75}{±6.71} \\

SOLT-GNN~\cite{soltgnn2022} & 
65.69\scalebox{0.75}{±2.36} & 
62.78\scalebox{0.75}{±1.33} & 
63.21\scalebox{0.75}{±3.40} \\

TopoImb~\cite{topoimb2022} & 
56.18\scalebox{0.75}{±2.10} & 
64.73\scalebox{0.75}{±7.09} & 
69.66\scalebox{0.75}{±1.92} \\
\midrule
SamGoG (w/ TailGNN~\cite{tailgnn}) &
\cellcolor{firstplace}{\bf\color{red}71.06\scalebox{0.75}{±3.54}} & 
\cellcolor{firstplace}{\bf\color{red}72.80\scalebox{0.75}{±2.53}} & 
\cellcolor{firstplace}{\bf\color{red}74.65\scalebox{0.75}{±1.45}} \\
SamGoG (w/ GraphSHA~\cite{graphsha}) &
\cellcolor{secondplace}{68.54}\scalebox{0.75}{±1.72} & 
\cellcolor{secondplace}{72.21}\scalebox{0.75}{±2.79} & 
\cellcolor{secondplace}{71.98}\scalebox{0.75}{±2.06} \\
\bottomrule
\end{tabular}
}
\label{tab:res_dd_size_gcn_acc}%
\end{table}

\end{document}